\documentclass{article}

\usepackage[final,nonatbib]{neurips_data_2022}

\usepackage[utf8]{inputenc} 
\usepackage[T1]{fontenc}    
\usepackage{hyperref}       
\usepackage{url}            
\usepackage{graphicx}       
\usepackage{subcaption}     
\usepackage{wrapfig}        
\usepackage{booktabs}       
\usepackage{amsfonts}       
\usepackage{nicefrac}       
\usepackage{microtype}      
\usepackage{xcolor}         
\usepackage{xspace}
\usepackage{makecell}
\usepackage{enumitem}       
\usepackage{pifont}
\usepackage{lipsum}
\usepackage[perpage]{footmisc}
\usepackage{listings}
\usepackage{multirow}
\usepackage{amsmath}

\definecolor{dkred}{rgb}{0.5,0,0}
\definecolor{dkgreen}{rgb}{0,0.6,0}
\definecolor{gray}{rgb}{0.5,0.5,0.5}
\definecolor{mauve}{rgb}{0.58,0,0.82}

\newcommand{\method}{{BOND}\xspace}
\newcommand{\prob}{{OND}\xspace}
\newcommand{\cmark}{\ding{51}}%
\newcommand{\xmark}{\ding{55}}%
\newtheorem{definition}{Definition}

\usepackage{soul}
\usepackage{blindtext}
\usepackage{todonotes}
\newcommand{\rv}[1]{{#1}}

\usepackage{colortbl}
\usepackage{soul}
\newcolumntype{y}{>{\columncolor{yellow}}c}

\title{
BOND: Benchmarking Unsupervised Outlier Node Detection on Static Attributed Graphs
}

\author{%
  Kay Liu$^{1,\ast}$, Yingtong Dou$^{1,8,\ast}$, Yue Zhao$^{2,}$\thanks{Equal Contribution.}, $\;$Xueying Ding$^{2}$,  \textbf{Xiyang Hu}$^{2}$, \\
  \textbf{Ruitong Zhang}$^{3}$,
  \textbf{Kaize Ding}$^{4}$,
  \textbf{Canyu Chen}$^{5}$, \textbf{Hao Peng}$^{3}$,  \textbf{Kai Shu}$^{5}$,\\
    \textbf{Lichao Sun}$^{6}$,
  \textbf{Jundong Li}$^{7}$,
  \textbf{George H. Chen}$^{2}$,
  \textbf{Zhihao Jia}$^{2}$,
  \textbf{Philip S. Yu}$^{1}$\\
  $^1$University of Illinois Chicago $^2$Carnegie Mellon University \\
  $^3$Beihang University $^4$Arizona State University $^5$Illinois Institute of Technology \\
  $^6$Lehigh University $^7$University of Virginia $^8$Visa Research\\
  \texttt{\href{mailto:benchmark@pygod.org}{benchmark@pygod.org}}\vspace{-1em}
  }

\begin{document}

\maketitle

\begin{abstract}\vspace{-1em}
Detecting which nodes in graphs are outliers is a relatively new machine learning task with numerous applications. 
Despite the proliferation of algorithms developed in recent years for this task, there has been no standard comprehensive setting for performance evaluation. Consequently, it has been difficult to understand which methods work well and when under a broad range of settings.
To bridge this gap, we present---to the best of our knowledge---the first comprehensive \underline{b}enchmark for unsupervised \underline{o}utlier \underline{n}ode \underline{d}etection on \rv{static attributed graphs}
called \method, with the following highlights.
(1) We benchmark the outlier detection performance of 14 methods ranging from classical matrix factorization to the latest graph neural networks.
(2) Using nine real datasets, our benchmark assesses how the different detection methods respond to two major types of synthetic outliers and separately to ``organic'' (real non-synthetic) outliers.
(3) Using an existing random graph generation technique, we produce a family of synthetically generated datasets of different graph sizes that enable us to compare the running time and memory usage of the different outlier detection algorithms.
Based on our experimental results, we discuss the pros and cons of existing graph outlier detection algorithms, and we highlight opportunities for future research.
Importantly, our code is freely available and meant to be easily extendable: \\ \url{https://github.com/pygod-team/pygod/tree/main/benchmark}


\end{abstract}

\section{Introduction}
\label{sec01:intro}
\vspace{-0.1in}

Outlier detection (OD) on a graph refers to the task of identifying which nodes in the graph are outliers. This is a key machine learning (ML) problem that arises in many applications, such as social network spammer detection~\cite{ye2015discovering}, 
sensor fault detection~\cite{gaddam2020detecting}, 
financial fraudster identification~\cite{dou2020enhancing}, and defense against graph adversarial attacks~\cite{ioannidis2021unveiling}. 
Unlike classical OD on tabular and time-series data, graph OD has additional challenges: (1) the graph data structure in general carries richer information, and thus more powerful ML models are needed to learn informative representations, and (2) 
with more complex ML models, training can be more computationally expensive in terms of both running time and memory consumption  \cite{jia2020improving,lai2020policy}, posing challenges for time-critical  (i.e., low time budget) and resource-sensitive (e.g., limited GPU memory) applications.

Despite the importance of graph OD and many algorithms being developed for it in recent years, \textit{there is no comprehensive benchmark on graph outlier detection}, which we believe has hindered the development and understanding of graph OD algorithms. In fact, a recent graph OD survey calls for ``system benchmarking'' and describes it as ``the key to evaluating the performance of graph OD techniques'' \cite{ma2021comprehensive}. We remark that there already are benchmarks for general graph mining (e.g.,  OGB \cite{ogb}), graph representation learning \cite{freitas2021large}, graph robustness evaluation \cite{zheng2021graph}, graph contrastive learning~\cite{zhu2021empirical}, graph-level anomaly detection \cite{zhao2021using},\footnote{Graph-level anomaly detection refers to when we have a set of graphs and want to find which graphs are significantly different from the majority of graphs; in contrast, the graph OD we focus on in this paper is for detecting outliers at the node level for a specific graph.} as well as benchmarks for tabular OD \cite{campos2016evaluation} and time-series OD~\cite{lai2021revisiting}. These do not cover the specific task we consider, which we now formally define:

\begin{definition}\vspace{-0.5em}
\label{def:problem-statement}
(Unsupervised \underline{O}utlier \underline{N}ode \underline{D}etection on Static Attributed Graphs (abbreviated as \prob))
A static attributed graph is defined as $G=(V, E, \mathbf{X})$, where $V=\{1,2,\dots,N\}$ is the set of vertices, $E\subseteq\{(i,j):i,j\in V\text{ s.t.~}i\ne j\}$ is the set of edges, and $\mathbf{X}\in\mathbb{R}^{N\times D}$ is the node attribute matrix (the $i$-th row of $\mathbf{X}$ is the feature vector in $\mathbb{R}^D$ corresponding to the $i$-th node in the graph).
Given the graph $G$, the goal of the problem \prob is to learn a function $f:V\rightarrow\mathbb{R}$ that assigns a real-valued outlier score to every node in $G$. The outlier nodes are then taken to be the $k$ nodes with the highest outlier scores, for a user-specified value of $k$.
This problem is unsupervised since in learning $f$, we do not have any ground truth information as to which nodes are outliers or not.
\end{definition}\vspace{-0.5em}

Note that there are other graph OD problems (e.g., feature vectors could be time-dependent, there could be supervision in terms of some outliers being labeled, the nodes and edges could change over time, etc) but we focus on the problem \prob stated above as it is the most prevalent \cite{ding2019deep,ma2021comprehensive}. For \prob,
the status quo for how algorithms are developed has the following limitations:
\setlist{nolistsep}
\begin{itemize}[leftmargin=*,noitemsep]
\item \textbf{Lack of a comprehensive benchmark:} often, only a limited selection of \prob algorithms is tested on only a few datasets, making it unclear to what extent the empirical results generalize to a wider range of settings. Here, we remark that this issue of generalization is exacerbated by the fact that across different applications, what constitutes an outlier can vary drastically and, at the same time, also be difficult to precisely define in a manner that domain experts agree upon.
\item \textbf{Limited outlier types taken into account:} typically, only a few types of outliers are considered (e.g., specific kinds of synthetic outliers are injected into real datasets), making it difficult to understand how graph OD algorithms respond to a wider variety of outlier nodes, including ones that are ``organic'' (non-synthetic).
\item 
\textbf{Limited analyses of computational efficiency in both time and space:}
Existing work mainly focuses on detection accuracy, with limited analyses of running time and memory consumption.
\end{itemize}
To address all the above limitations, we establish \textit{the first comprehensive benchmark for the problem of \prob} that we call \method (short for \underline{b}enchmarking unsupervised \underline{o}utlier \underline{n}ode \underline{d}etection on static attributed graphs). To accommodate many algorithms, we specifically create an open-source \underline{Py}thon library for \underline{G}raph \underline{O}utlier \underline{D}etection (PyGOD)\footnote{A \underline{Py}thon library for \underline{G}raph \underline{O}utlier \underline{D}etection (PyGOD): \url{https://pygod.org/}},
which provides more than ten of the latest graph OD algorithms, all with unified APIs and optimizations. Meanwhile, PyGOD also includes multiple non-graph baselines, resulting in a total of 14 representative and diverse methods for \prob. We remark that this library can readily be extended to include additional OD algorithms.

Our work has the following highlights:
\setlist{nolistsep}
\begin{enumerate}[leftmargin=*,noitemsep]
    \item \textbf{The first comprehensive node-level graph OD benchmark}. We examine 14 OD methods, including classical and deep ones, and compare their pros and cons on nine benchmark datasets.
    \item \textbf{Consolidated taxonomy of outlier nodes.} We group existing notions of outlier nodes into two main types: structural and contextual outliers. Our results show that most methods fail to balance the OD performance of these two major outlier types.
    \item \textbf{Systematic performance flaw found for existing deep graph OD methods.} Surprisingly, our experimental results in \method reveal that most of the benchmarked deep graph OD methods have suboptimal OD performance on organic outliers.
    \item \textbf{Evaluation of both detection quality and computational efficiency}. In addition to common \textit{effectiveness} metrics (e.g., ROC-AUC), we also measure the running time and GPU memory consumption of different algorithms as their \textit{efficiency} measures.
    \item \textbf{Reproducible and accessible benchmark toolkit}. To foster accessibility and fair evaluation for future algorithms, we make our code for \method freely available at: \\ \url{https://github.com/pygod-team/pygod/tree/main/benchmark}
\end{enumerate}
We briefly describe existing approaches for \prob in \S \ref{sec02:rw}. We provide an overview of \method in \S \ref{sec03:gob}, followed by detailed experimental results and analyses in \S \ref{sec04:exp}. We summarize the paper and discuss future work in \S \ref{sec06:con}.
\vspace{-0.1in}

\section{Related Work}
\label{sec02:rw}
\vspace{-0.1in}

In this section, we briefly introduce related work on outlier node detection.
Please refer to~\cite{akoglu2015graph} and~\cite{ma2021comprehensive} for more comprehensive reviews of classical and deep-learning-based graph outlier detectors. We have implemented most of the discussed algorithms in this section in the PyGOD.

\textbf{Classical (non-deep) outlier node detection.}
Real-world evidence suggests that the outlier nodes are different from regular nodes in terms of structure or attributes.
Thus, early work on node outlier detection employs graph-based features such as centrality measures and clustering coefficients to extract the anomalous signals from graphs~\cite{akoglu2015graph}.
Instead of handcrafting features, learning-based methods have been used to more flexibly encode graph information to spot outlier nodes. Examples of these learning-based methods include ones based on matrix factorization (MF)~\cite{bandyopadhyay2019outlier,li2017radar, peng2018anomalous,tong2011non}, density-based clustering~\cite{chakrabarti2004autopart,hooi2016fraudar,xu2007scan}, and relational learning~\cite{koutra2011unifying,rayana2015collective}.
As most of the methods above have constraints on graph/node types or prior knowledge, we only include SCAN~\cite{xu2007scan}, Radar~\cite{li2017radar} and ANOMALOUS~\cite{peng2018anomalous} in \method to represent methods in this category.

\textbf{Deep outlier node detection.}
The rapid development of deep learning and its use with graph data has shifted the landscape of outlier node detection from traditional methods to neural network approaches~\cite{ma2021comprehensive}. For example,
the autoencoder (AE)~\cite{kingma2013auto}, which is a neural network architecture devised to learn an encoding of the original data by trying to reconstruct the original data from the encoding, has become a popular model in detecting outlier nodes~\cite{ding2019deep,fan2020anomalydae,kipf2016variational,sakurada2014anomaly, wang2018deep,yuan2021higher}.
AEs can be learned in an unsupervised manner as we are aiming to reconstruct the original data without separately trying to predict labels.
The heuristic behind AE-based outlier detection is that we can use the AE reconstruction error as an outlier score; a data point that has a higher reconstruction error is likely more atypical.

More recently, graph neural networks (GNNs) have attained superior performance in many graph mining tasks~\cite{ding2022meta,ding2022data,kipf2016semi,wang2020next,zha2022towards}. GNNs aim to learn an encoding representation for every node in the graph, taking into account node attributes and also the underlying graph structure. The encoding representations learned by GNNs turn out to capture complex patterns that are useful for OD. As a result, GNNs have also become popular in detecting outlier nodes in graphs~\cite{ding2021few,wang2021one, zhao2021synergistic,liu2021anomaly,xu2022contrastive}.
Note that GNNs can be combined with AEs; in constructing an AE, we need to specify encoder and decoder networks, which could be set to be GNNs.

We point out that it is also possible to adopt a Generative Adversarial Network (GAN) for outlier node detection~\cite{chen2020generative}. GANs learn how to generate fake data that resemble real data by simultaneously learning a generator network (that can be used to randomly generate fake data) and a discriminator network (that tries to tell whether a data point is real or fake). Naturally, outliers could be deemed to be data points that are considered more ``fake''.

Among the many deep outlier node detectors, we have thus far implemented nine (see Table~\ref{table:algorithms}) for inclusion in \method, where we have tried to have these be somewhat diverse in their methodology.

\vspace{-0.1in}

\section{BOND}
\label{sec03:gob}
\vspace{-0.1in}

In this section, we provide an overview of \method. We begin by defining two outlier types in \S \ref{subsec:taxonomy}. We then elaborate on the datasets (\S \ref{subsec:datasets}), algorithms (\S \ref{subsec:algorithms}), and evaluation metrics (\S \ref{subsec:metrics}) in \method.

\vspace{-0.125in}
\subsection{Outlier Types}
\label{subsec:taxonomy}
\vspace{-0.1in}

\begin{wrapfigure}{r}{0.45\textwidth}
\vspace{-1em}
  \includegraphics[width=0.45\textwidth]{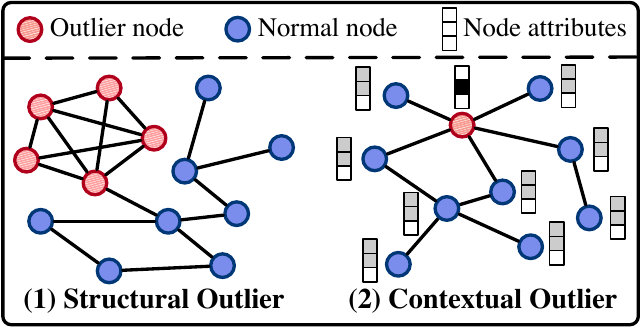}
    \vspace{-0.15in}
  \caption{An illustration of structural vs. contextual outliers.}
  \vspace{-0.3in}
  \label{fig:outlier_type}
\end{wrapfigure}

Many researchers have defined fine-grained outlier node types from different perspectives ~\cite{akoglu2015graph,bandyopadhyay2019outlier,ding2019deep,ioannidis2021unveiling,li2017radar,ma2021comprehensive}.
In this paper, we group existing outlier node definitions into two major types according to real-world outlier patterns: \textit{structural outliers} and \textit{contextual outliers}, which are illustrated in Figure~\ref{fig:outlier_type} and defined below.

\vspace{-.5em}
\begin{definition}(Structural outlier)
\label{def:structural-outlier}
Structural outliers are densely connected nodes in contrast to sparsely connected regular nodes.
\end{definition}
\vspace{-.5em}

Structural outliers arise in many real-world applications. For example, members of organized fraud gangs who frequently collude in carrying out malicious activities can be viewed as forming dense subgraphs of an overall graph (with nodes representing different people)~\cite{akoglu2015graph}.
As another example, coordinated bot accounts retweeting the same tweet will form a densely-connected co-retweet graph~\cite{hooi2016fraudar, pacheco2021uncovering}. 
Note that some papers~\cite{li2017radar,ma2021comprehensive} also regard isolated nodes that do not belong to any communities as structural outliers (i.e., they have only a few edges connecting to any communities), which is different from our definition above.
Since there is no existing OD method that we are aware of for detecting these isolated outlier nodes, we do not cover this type of outlier in \method.

\vspace{-.5em}
\begin{definition}(Contextual outlier)
\label{def:contextual-outlier}
Contextual outliers are nodes whose attributes are significantly different from their neighboring nodes.
\end{definition}
\vspace{-.5em}

An example of a contextual outlier is a compromised device in a computer network~\cite{akoglu2015graph}.
The definition of a contextual outlier is similar to how outliers are defined in classical proximity-based OD methods~\cite{aggarwal2017introduction}.

Some researchers call a node whose attributes differ from those of all other nodes as a \emph{global outlier}~\cite{ma2021comprehensive} or a contextual outlier~\cite{li2017radar}. We do not consider these outliers in \method as we find that they do not actually use the graph structure; instead, these outliers could be detected using tabular outlier detectors~\cite{han2022adbench,zhao2019pyod}.

What we defined as a contextual outlier in Definition~\ref{def:contextual-outlier} is also referred to as an attribute outlier~\cite{ding2019deep} or a community outlier~\cite{li2017radar, ma2021comprehensive} in previous work.
We argue that calling these \textit{contextual outliers} is a more accurate terminology. The reason is that an ``attribute outlier'' sounds like it only depends on attributes (i.e., feature vectors), which would correspond to a global outlier~\cite{ma2021comprehensive}, but confusingly this is not what is meant by the terminology. Meanwhile, the terminology of a ``community'' in graph theory has often been in reference to the density of edges among nodes and so a ``community outlier'' might be misconstrued to be what we call a \emph{structural outlier} as in Definition~\ref{def:structural-outlier}.
For the remainder of the paper, by ``structural'' and ``contextual'' outliers, we always go by Definitions~\ref{def:structural-outlier} and \ref{def:contextual-outlier} respectively.

Importantly, note that in real datasets, the organic (non-synthetic) outlier nodes present do not need to strictly be either a structural or a contextual outlier. In fact, what precisely makes them an outlier need not be explicitly stated and they could be neither a structural nor a contextual outlier, or they could even appear as a mixture of these two types! This makes detecting organic outliers more difficult than detecting synthetic outliers that follow a specific pattern such as those of Definitions~\ref{def:structural-outlier} and \ref{def:contextual-outlier}.

\vspace{-0.125in}
\subsection{Datasets}
\label{subsec:datasets}
\vspace{-0.1in}

To comprehensively evaluate the performance of existing \prob algorithms, \rv{we have investigated various real datasets with organic outliers used in previous literature. Note that some standard datasets are beyond the scope of the problem \prob that we consider or do not make use of either the graph structure or node attributes/feature vectors. For example, YelpChi-Fraud~\cite{dou2020enhancing}, Amazon-Fraud~\cite{dou2020enhancing}, and Elliptic~\cite{weber2019anti} are three graph datasets designed for supervised node classification; however, the fraudulent nodes have limited outlier pattern in terms of graph structure.
Bitcoin-OTC, Bitcoin-Alpha, Epinions, and Amazon-Malicious from~\cite{kumar2018rev2} are four bipartite graphs where nodes do not have attributes. DARPA~\cite{yoon2019fast}, UCI Message~\cite{zheng2019addgraph}, and Digg~\cite{zheng2019addgraph} are three dynamic graphs with organic edge outliers which are also out of our problem scope.
}

In \method, we use the following datasets. First, since there are a limited number of open-source graph datasets with organic outlier nodes, we include three real datasets with no organic outliers that we inject synthetic outlier nodes into. Specifically, we use node classification benchmark datasets (\textbf{Cora}~\cite{sen2008collective}, \textbf{Amazon}~\cite{shchur2018pitfalls}, and \textbf{Flickr}~\cite{zeng2019graphsaint}) from three domains with different scales. Next, we use six real datasets that contain organic outliers (\textbf{Weibo}~\cite{zhao2020error}, \textbf{Reddit}~\cite{kumar2019predicting, wang2021decoupling}, \rv{\textbf{Disney}~\cite{sanchez2013statistical},
\textbf{Books}~\cite{sanchez2013statistical},
\textbf{Enron}~\cite{sanchez2013statistical}, and
\textbf{DGraph}~\cite{huang2022dgraph}}). Finally, we also use purely synthetic data generated using the random algorithm by~\cite{kim2022find} that is able to produce graphs with varying scales; this random generation procedure provides a controlled manner in which we can evaluate different OD algorithms' computational efficiency in terms of both running time and memory usage. Some basic statistics for the real datasets used are given in Table~\ref{tab:data-stat} with more dataset details available in Appx.~\ref{appdx-dataset}.

\begin{table}[!ht]
\centering
\caption{Statistics of real datasets used in \method ($^*$ indicates that outliers are synthetically injected).}
\label{tab:data-stat}
\scalebox{0.92}{
\begin{tabular}{@{}l|cccccccc@{}}
\toprule
\textbf{Dataset} & \textbf{\#Nodes} & \textbf{\#Edges} & \textbf{\#Feat.   } & \textbf{Degree} & \textbf{\#Con.} & \textbf{\#Strct.} & \textbf{\#Outliers} & \textbf{Ratio}  \\ \midrule
\textbf{Cora$^{*}$~\cite{sen2008collective}}   & 2,708            & 11,060           & 1,433               & 4.1             & 70              & 70                       & 138            & 5.1\%            \\
\textbf{Amazon$^{*}$~\cite{shchur2018pitfalls}} & 13,752           & 515,042          & 767                 & 37.2            & 350             & 350                      & 694            & 5.0\%            \\
\textbf{Flickr$^{*}$~\cite{zeng2019graphsaint}} & 89,250           & 933,804          & 500                 & 10.5            & 2,240           & 2,240                    & 4,414          & 4.9\%            \\
\textbf{Weibo~\cite{zhao2020error}}        & 8,405            & 407,963          & 400                 & 48.5            & -               & -                        & 868            & 10.3\%           \\
\textbf{Reddit~\cite{kumar2019predicting, wang2021decoupling}}       & 10,984           & 168,016          & 64                  & 15.3            & -               & -                        & 366            & 3.3\%            \\
\textbf{Disney~\cite{sanchez2013statistical}}       & 124           & 335         & 28                & 2.7           & -               & -                        & 6            & 4.8\%           \\
\textbf{Books~\cite{sanchez2013statistical}}       & 1,418         & 3,695       & 21                  & 2.6            & -               & -                        & 28            & 2.0\%            \\
\textbf{Enron~\cite{sanchez2013statistical}}       & 13,533           & 176,987         & 18                & 13.1           & -               & -                        & 5            & 0.4\textperthousand          \\
\textbf{DGraph~\cite{huang2022dgraph}}       & 3,700,550           & 4,300,999          & 17                 & 1.2            & -               & -                        & 15,509            & 0.4\%            \\
\bottomrule
\end{tabular}}
\vspace{-1.em}
\end{table}

To make synthetic outlier nodes of the two types we defined in \S \ref{subsec:taxonomy} and to ``camouflage'' them so that they are more difficult to detect using simple OD methods, we slightly modify a widely-used approach~\cite{ding2019deep, fan2020anomalydae, chen2020generative, yuan2021higher} (described below). These synthetic outliers are used with the real datasets that lack organic outliers (Cora, Amazon, Flickr) and also with the randomly generated graph data. In the random outlier injection procedures to follow, for the given graph $G$ that we are working with, we treat the vertex set as fixed. To inject structural outliers, we modify the edges present, whereas to inject contextual outliers, we modify the feature vectors of randomly chosen nodes.

\textbf{Injecting random structural outliers.}
The basic strategy is to create $n$ non-overlapping densely connected groups of nodes, where each group has exactly $m$ nodes (so that there are a total of $m\times n$ structural outliers injected). To do this, for each $i = 1, \dots, n$, we randomly sample $m$ nodes to form the $i$-th group (these $m$ nodes are sampled uniformly at random from nodes that have not been previously chosen to form a group); for these $m$ nodes, we first make them fully connected and then drop each edge independently with probability $p$. 

\textbf{Injecting random contextual outliers.}
To inject a total of $o$ contextual outliers, we first sample $o$ nodes from the vertex set $V$ without replacement; these are the nodes whose attributes we aim to modify as to turn them into contextual outliers. We denote the set of these $o$ nodes as $V_c$ (so that $o=|V_c|$), and refer to the remaining nodes $V_r:=V \setminus V_c$ as the ``reference'' set. For each node $i \in V_c$, we randomly choose $q$ nodes without replacement uniformly at random from the reference set $V_r$. Among these $q$ reference nodes chosen, we find the one whose attributes deviate the most (in terms of Euclidean distance) from those of node $i$. We then change the attributes of node $i$ to be the same as those of this most dissimilar reference node found.

For more details about synthetic outlier injection, see Appx~\ref{outlier-inject}.




\vspace{-0.125in}
\subsection{Algorithms}
\label{subsec:algorithms}
\vspace{-0.1in}

\begin{table}[!t]
\caption{Algorithms implemented in \method and their characteristics: whether designed for graphs (row 3), whether neural networks are used (row 4), and what the core idea for the method (row~5).} 
\label{table:algorithms}
\centering
    \scalebox{0.61}{
	\begin{tabular}{l|ccccccccccccc }
	\toprule
        \textbf{Alg.} & \textbf{LOF} & \textbf{IF} & \textbf{MLPAE} & \textbf{SCAN}  & \textbf{Radar} & \textbf{\makecell{ANOMA-\\LOUS}} & \textbf{GCNAE} & \textbf{\makecell{DOMI-\\NANT}} & \textbf{\makecell{DONE/\\AdONE}} & \textbf{\makecell{Anomaly-\\DAE}} & \textbf{GAAN} & \textbf{GUIDE} & \textbf{CONAD} \\ \midrule
        \textbf{Year}      & 2000         & 2012        & 2014      & 2007           & 2017           & 2018               & 2016           & 2019              & 2020                         & 2020                & 2020          & 2021           & 2022           \\
        \textbf{Graph}     & \xmark       & \xmark      & \xmark    & \cmark         & \cmark         & \cmark             & \cmark         & \cmark         & \cmark            & \cmark                       & \cmark        & \cmark          & \cmark        \\
        \textbf{Deep}      & \xmark       & \xmark      & \cmark    & \xmark         & \xmark         & \xmark             & \cmark         & \cmark         & \cmark            & \cmark                       & \cmark        & \cmark          & \cmark        \\
        \textbf{Core}  & N/A          & Tree         & MLP+AE            & Cluster             & MF             & MF                 & GNN+AE            & GNN+AE               & MLP+AE          & GNN+AE                 & GAN           & GNN+AE            & GNN+AE           \\
        \textbf{Ref.} & \cite{breunig2000lof} & \cite{liu2012isolation} & \cite{sakurada2014anomaly} & \cite{xu2007scan} & \cite{li2017radar} & \cite{peng2018anomalous} & \cite{kipf2016variational} & \cite{ding2019deep} & \cite{bandyopadhyay2020outlier} & \cite{fan2020anomalydae} & \cite{chen2020generative} & \cite{yuan2021higher} & \cite{xu2022contrastive} \\
    \bottomrule
	\end{tabular}}
\vspace{-1.5em}
\end{table}

Table~\ref{table:algorithms} lists the 14 algorithms evaluated in the benchmark and their properties. Our principle for selecting algorithms to implement in \method is to cover representative methods in terms of the published time (``Year''), whether they use the graph structure (``Graph''), whether they use neural networks (``Deep''), and what the core idea is behind the method (``Core'').
By including non-graph OD algorithms (LOF, IF, MLPAE), we can investigate the advantages and deficiencies of graph-based vs. non-graph-based OD algorithms in detecting outlier nodes.
\rv{Similarly, incorporating three classical OD methods (clustering-based SCAN, MF-based Radar, and ANOMALOUS) helps us understand the performance of classical vs. deep OD methods.
We select a wide array of GNN-based methods including the vanilla GCNAE; the classic DOMINANT; AnomalyDAE, an improved version of DOMINANT; and also GUIDE and CONAD, two state-of-the-art methods in this category with different data augmentation techniques.
Besides GNN-based \method methods, two methods encoding graph information using other models (DONE/AdONE and GAAN) are also included.}
Please refer to Appx.~\ref{appdx-algorithms} for a more detailed introduction of the methods benchmarked in \method.

\vspace{-0.125in}
\subsection{Evaluation Metrics}
\label{subsec:metrics}
\vspace{-0.1in}

\textbf{Detection quality measures}. \rv{We follow the extensive literature in graph OD  \cite{ding2021inductive,tong2011non,zhao2021synergistic} to comprehensively evaluate the outlier node detection quality with three metrics: (1) ROC-AUC reflects detectors' performance on both positive and negative examples, while (2) Average Precision focuses more on positive examples, and (3) Recall@k evaluates the examples with high predicted outlier scores.} See Appx.~\ref{appdx-metrics} for more details.

\textbf{Efficiency measures in time and space}. Another important aspect of graph-based algorithms is their high time and space complexity \cite{ding2021vq,jia2020improving}, which imposes additional challenges for large, high-dimensional datasets on hardware like GPUs with limited memory (e.g., out-of-memory errors). Therefore, we measure efficiency in time and space respectively, we use (1) wall-clock time and (2) GPU memory consumption. We provide more details in Appx.~\ref{appdx-metrics}.

\vspace{-0.1in}

\section{Experiments}
\label{sec04:exp}
\vspace{-0.1in}

We design \method to understand the detection effectiveness and efficiency of various OD algorithms in addressing the problem \prob.  
Specifically, we aim to answer:
\rv{\textbf{RQ1} (\S \ref{subsub:detection-perform}): How effective are the algorithms on detecting synthetic and organic outliers?
\textbf{RQ2} (\S \ref{subsub:outlier_types}): How do algorithms perform under two types of synthetic outliers (structural and contextual)?
\textbf{RQ3} (\S \ref{subsub:efficiency}): 
How efficient are algorithms in terms of time and space? 
Note that due to space constraints, for detection quality, we focus on the ROC-AUC metric, deferring results using the AP and Recall@k metrics to Appx.~\ref{appedix:add_results}}.


\textbf{Model implementation and environment configuration.}
Most algorithms in \method are implemented via our newly released PyGOD package~\cite{liu2022pygod}, and non-graph OD methods are imported from our earlier work~\cite{zhao2019pyod}.
Although we tried our best to apply the same set of optimization techniques, e.g., vectorization, to all methods, we suspect that further code optimization is possible.
For more implementation details and environment configurations, see Appx.~\ref{appedix:exp_setting}.



\textbf{Hyperparameter grid.} In real-world settings, it is unclear how to do hyperparameter tuning and algorithm selection for unsupervised outlier detection due to the lack of ground truth labels and/or universal criteria that correlates well with the ground truth \cite{ma2021comprehensive,zhao2021automatic}. For fair evaluation, when we report performance metrics in tables, we apply the same hyperparameter grid (see Appx.~\ref{appedix:exp_setting}) to each applicable algorithm and report its \textit{avg. performance} (i.e., ``algorithm performance in expectation''), along with the \textit{standard deviation} (i.e., ``algorithm stability'') and the \textit{max} (i.e., ``algorithm potential'').



\vspace{-0.125in}
\subsection{Experimental Results: Detection Performance on Synthetic and Organic Outliers}
\label{subsub:detection-perform}
\vspace{-0.1in}

Using the nine real datasets described in \S \ref{subsec:datasets}, we report the ROC-AUC score of different OD algorithms in Tables \ref{tab:main-table} and \ref{tab:main-table-real}. Below are the key findings from these tables.

\begin{table}[t]
\centering
\caption{\small ROC-AUC (\%) comparison among OD algorithms on three datasets with synthetic outliers, where we show \textit{the avg perf.} $\pm$ \textit{the STD of perf.} (\textit{max perf.}) of each. The best algorithm by expectation is shown in \textbf{bold}, while the max performance per dataset is marked with \underline{underline}. OOM denotes out of memory with regard to GPU (\_G) and CPU (\_C).}
\label{tab:main-table}
\scalebox{0.71}{
\begin{tabular}{@{}l|ccc@{}}
\toprule
\textbf{Algorithm}  & \textbf{Cora}         & \textbf{Amazon}       & \textbf{Flickr}  \\ \midrule
\textbf{LOF}        & 69.9±0.0 (69.9) & 55.2±0.0 (55.2)  & 41.6±0.0 (41.6)   \\
\textbf{IF}         & 64.4±1.5 (67.4) & 51.3±3.0 (57.9)  & 57.1±1.1 (58.8)   \\
\textbf{MLPAE}      & 70.9±0.0 (70.9) & 74.2±0.0 (74.2)  & 72.4±0.0 (72.5)  \\
\midrule
\textbf{SCAN}       & 62.8±4.5 (72.6)  & 62.2±4.9 (71.1) & 62.4±12.4 (75.0) \\
\textbf{Radar}      & 65.0±1.3 (66.0)  & 71.8±1.1 (73.4) & OOM\_G           \\
\textbf{ANOMALOUS}  & 55.0±10.3 (68.0) & 72.5±1.5 (75.5) & OOM\_G           \\
\midrule
\textbf{GCNAE}      & 70.9±0.0 (70.9)  & 74.2±0.0 (74.2) & 71.6±3.1 (72.4)  \\
\textbf{DOMINANT}   & 82.7±5.6 (84.3)  & 81.3±1.0 (82.2) & 78.0±12.0 (84.6) \\
\textbf{DONE}       & 82.4±5.6 (\underline{87.9})  & 82.8±8.8 (\underline{93.7}) & \textbf{84.7}±2.5 (\underline{89.0})  \\
\textbf{AdONE}      & 81.5±4.5 (87.4)  & \textbf{86.6}±5.6 (92.3) & 82.8±3.2 (\underline{89.0})  \\
\textbf{AnomalyDAE} & \textbf{83.4}±2.3 (85.3)  & 85.7±2.9 (90.8) & 65.6±3.5 (70.4)  \\
\textbf{GAAN}       & 74.2±0.9 (76.1)  & 80.8±0.3 (81.5) & 72.4±0.2 (72.5)  \\
\textbf{GUIDE}      & 74.7±1.3 (77.5)  & OOM\_C          & OOM\_C           \\
\textbf{CONAD}      & 78.8±9.6 (84.3)  & 80.5±4.0 (82.2) & 65.1±2.5 (67.4)  \\
\bottomrule
\end{tabular}}
\vspace{-1.4em}
\end{table}
\begin{table}[t]
\centering
\caption{\small ROC-AUC (\%) comparison among OD algorithms on six datasets with organic outliers, where we show \textit{the avg perf.} $\pm$ \textit{the STD of perf.} (\textit{max perf.}) of each. The best algorithm by expectation is shown in \textbf{bold}, while the max performance per dataset is marked with \underline{underline}. OOM denotes out of memory with regard to GPU (\_G) and CPU (\_C). TLE denotes time limit of 24 hours exceeded.}
\label{tab:main-table-real}
\scalebox{0.71}{
\begin{tabular}{@{}l|cc cccc@{}}
\toprule
\textbf{Algorithm} & \textbf{Weibo}  & \textbf{Reddit}   & \textbf{Disney}   & \textbf{Books}       & \textbf{Enron}     & \textbf{DGraph}      \\ 
\midrule
\textbf{LOF}  & 56.5±0.0 (56.5)  & \textbf{57.2}±0.0 (57.2) & 47.9±0.0 (47.9) & 36.5±0.0 (36.5) & 46.4±0.0 (46.4)  & TLE     \\
\textbf{IF}   & 53.5±2.8 (57.5)  & 45.2±1.7 (47.5)  & \textbf{57.6}±2.9 (63.1) & 43.0±1.8 (47.5) & 40.1±1.4 (43.1)  & \textbf{60.9}±0.7 (\underline{62.0}) \\
\textbf{MLPAE}      & 82.1±3.6 (86.1)  & 50.6±0.0 (50.6)  & 49.2±5.7 (\underline{64.1}) & 42.5±5.6 (52.6) & 73.1±0.0 (73.1)  & 37.0±1.9 (41.3) \\
\midrule
\textbf{SCAN}       & 63.7±5.6 (70.8) & 49.9±0.3 (50.0) & 50.5±4.0 (56.1) & 49.8±1.7 (52.4) & 52.8±3.4 (58.1)  & TLE             \\
\textbf{Radar}      & \textbf{98.9}±0.1 (\underline{99.0})  & 54.9±1.2 (56.9) & 51.8±0.0 (51.8) & 52.8±0.0 (52.8) & \textbf{80.8}±0.0 (80.8)  & OOM\_C          \\
\textbf{ANOMALOUS}  & \textbf{98.9}±0.1 (\underline{99.0})  & 54.9±5.6 (\underline{60.4}) & 51.8±0.0 (51.8) & 52.8±0.0 (52.8) & \textbf{80.8}±0.0 (80.8)  & OOM\_C          \\
\midrule
\textbf{GCNAE}      & 90.8±1.2 (92.5)  & 50.6±0.0 (50.6) & 42.2±7.9 (52.7) & 50.0±4.5 (57.9) & 66.6±7.8 (80.1)  & 40.9±0.5 (42.2) \\
\textbf{DOMINANT}   & 85.0±14.6 (92.5) & 56.0±0.2 (56.4) & 47.1±4.5 (54.9) & 50.1±5.0 (58.1) & 73.1±8.9 (\underline{85.0})  & OOM\_C          \\
\textbf{DONE}       & 85.3±4.1 (88.7)  & 53.9±2.9 (59.7) & 41.7±6.2 (50.6) & 43.2±4.0 (52.6) & 46.7±6.1 (67.1)  & OOM\_C          \\
\textbf{AdONE}      & 84.6±2.2 (87.6)  & 50.4±4.5 (58.1) & 48.8±5.1 (59.2) & 53.6±2.0 (56.1) & 44.5±2.9 (53.6)  & OOM\_C          \\
\textbf{AnomalyDAE} & 91.5±1.2 (92.8)  & 55.7±0.4 (56.3) & 48.8±2.2 (55.4) & \textbf{62.2}±8.1 (\underline{73.2}) & 54.3±11.2 (69.1) & OOM\_C          \\
\textbf{GAAN}       & 92.5±0.0 (92.5)  & 55.4±0.4 (56.0) & 48.0±0.0 (48.0) & 54.9±5.0 (61.9) & 73.1±0.0 (73.1)  & OOM\_C          \\
\textbf{GUIDE}      & OOM\_C              & OOM\_C       & 38.8±8.9 (52.5) & 48.4±4.6 (63.5) & OOM\_C           & OOM\_C          \\
\textbf{CONAD}      & 85.4±14.3 (92.7) & 56.1±0.1 (56.4) & 48.0±3.5 (53.1) & 52.2±6.9 (62.9) & 71.9±4.9 (84.9)  & 34.7±1.2 (36.5) \\
\bottomrule
\end{tabular}}
\vspace{-1.4em}
\end{table}

\textbf{In terms of avg.~performance, no outlier node detection method is universally the best on all datasets.}
\rv{Tables \ref{tab:main-table} and \ref{tab:main-table-real} show that only three of 14 methods evaluated (AnomalyDAE, Radar, ANOMALOUS) have the best avg.~performance (for the ROC-AUC metric) on two datasets (Cora and Books for AnomalyDAE; Weibo and Enron for Radar and ANOMALOUS).
The classical methods Radar and ANOMALOUS both have the best performance on Weibo and Enron (see Table~\ref{tab:main-table-real}) but they are worse than many deep learning methods on detecting synthetic outliers (see Table~\ref{tab:main-table}).}
Additionally, there is a substantial performance gap between the best- and worst-performing algorithms, e.g., DONE achieves $2.06 \times$ higher average ROC-AUC compared to LOF on Flickr.

\rv{\textbf{Most methods evaluated fail to detect organic outliers.}
Since most methods we evaluated are designed to handle structural and contextual outliers as defined in \S \ref{subsec:taxonomy}, to figure out the reason behind the failure and success in detecting organic outliers, we analyze the organic outlier patterns in terms of metrics related to the definitions of structural and contextual outliers.
We first show that the success of most methods on Weibo (see Table \ref{tab:main-table-real}) is because the outliers in Weibo exhibit the properties of both structural and contextual outliers.
Specifically, in Weibo, the average clustering coefficient~\cite{fagiolo2007clustering} of the outliers is higher than that of inliers (0.400 vs.~0.301), meaning that these outliers correspond to structural outliers. Meanwhile, the average neighbor feature similarity~\cite{dou2020enhancing} of the outliers is far lower than that of inliers (0.004 vs.~0.993), so that the outliers also correspond to contextual outliers.
In contrast, the outliers in the Reddit and DGraph datasets have similar average neighbor feature similarities and clustering coefficients for outliers and inliers.
Therefore, their abnormalities rely more on outlier annotations with domain knowledge, and so supervised OD methods are more effective than unsupervised ones on Reddit (best AUC: 0.746 in \cite{wang2021decoupling} vs.~0.604 in Table \ref{tab:main-table-real}) and DGraph (best AUC: 0.792 in \cite{huang2022dgraph} vs.~0.620 in Table \ref{tab:main-table-real}) than unsupervised ones.}

\rv{\textbf{Deep learning methods and other methods using SGD may be sub-optimal on small graphs.}
The outliers on Disney, Books, and Enron also have similar outlier patterns defined in \S \ref{subsec:taxonomy}.
However, most of the deep learning methods evaluated do not work particularly well on Disney and Enron compared to classical baselines.
The reason is that Disney and Books have small graphs in terms of \#Nodes, \#Edges, and \#Feat.~(see Table \ref{tab:data-stat}).
The small amount of data could make it difficult for the deep learning methods to encode the inlier distribution well and could also possibly lead to overfitting issues. Meanwhile, classical methods Radar and ANOMALOUS also perform poorly on Disney and Books; these methods use SGD, which we suspect could be problematic for such small datasets.
}

\rv{\textbf{Different categories of methods evaluated are good at detecting different types of outliers.}
According to Tables \ref{tab:main-table} and \ref{tab:main-table-real}, many deep graph-based methods are good at detecting synthetic outliers but are useless in detecting organic outliers.
Meanwhile, non-graph-based methods have advantages when outliers do not follow taxonomies (Reddit and DGraph).
These observations corroborate our understanding of unsupervised OD algorithms---their effectiveness depends on whether the underlying data distribution satisfies structural properties that the algorithms exploit.}

\textbf{In terms of standard deviation of the ROC-AUC metric, among the deep learning methods, some are noticeably less stable than others.}\footnote{For a specific method, the standard deviation of the method's ROC-AUC scores across hyperparameters (what we have called ``algorithm stability'') and the maximum (``algorithm potential'') are in general monotonically related (i.e., when the standard deviation increases, then the maximum \emph{minus the mean} also tends to increase; and vice versa), so that our finding here is both for algorithm stability and potential.}
Certain deep learning methods, e.g., GAAN, exhibit insensitivity to hyperparameters, where the ROC-AUC standard deviation is mostly below $1\%$.
Meanwhile, the methods that are unstable tend to involve more complex loss terms (e.g., weighted combination of multiple losses); for instance, DONE and AdONE achieve the highest \emph{max} performance (i.e., the numbers in parentheses in Tables~\ref{tab:main-table} and \ref{tab:main-table-real}, and not the \emph{avg.}~performance) among deep graph methods on three datasets, while showing high ROC-AUC standard deviation across hyperparameters tested. Here, we emphasize that in practice for unsupervised OD, there being no labels means that hyperparameter tuning is far less straightforward, so stability (in OD detection quality) across hyperparameters is a desirable property of an algorithm.

\vspace{-0.125in}
\subsection{Experimental Results: Detection Performance on Structural and Contextual Outliers}
\label{subsub:outlier_types}
\vspace{-0.1in}

\begin{table}[t]
\centering
\caption{\small ROC-AUC (\%) comparison among OD algorithms on three datasets injected with contextual and structural outliers, where we show \textit{the avg perf.} $\pm$ \textit{the STD of perf.} (\textit{max perf.}) of each. The best algorithm by expectation is shown in \textbf{bold}, while the max performance per dataset is marked with \underline{underline}. OOM denotes out of memory with regard to GPU (\_G) and CPU (\_C). Reconstruction-based MLPAE and GCNAE perform best w.r.t contextual outliers, while there is no universal winner for both types of outliers.}
\label{tab:outlier-types}
\scalebox{0.71}{
\begin{tabular}{@{}l|cccccc@{}}
\toprule
\textbf{}           & \multicolumn{2}{c}{\textbf{Cora}}             & \multicolumn{2}{c}{\textbf{Amazon}}           & \multicolumn{2}{c}{\textbf{Flickr}}           \\
\cmidrule{2-7}
\textbf{Algorithm}           & Contextual            & Structural            & Contextual            & Structural            & Contextual            & Structural            \\ \midrule
\textbf{LOF}         & 87.1±0.0 (87.1)  & 52.4±0.0 (52.4)  & 61.9±0.0 (61.9)   & 48.0±0.0 (48.0)   & 34.2±0.0 (34.2)   & 49.1±0.0 (49.1)   \\
\textbf{IF}          & 77.5±2.2 (81.8)  & 51.4±2.3 (56.2)  & 51.8±6.0 (64.3)   & 50.9±0.8 (52.2)   & 63.1±2.1 (66.5)   & 50.9±0.3 (51.5)   \\
\textbf{MLPAE}       & \textbf{88.9}±0.0 (\underline{88.9})  & 52.5±0.0 (52.5)  & \textbf{98.6}±0.0 (\underline{98.6})   & 49.0±0.0 (49.0)   & \textbf{94.4}±0.1 (\underline{94.5})   & 50.0±0.1 (50.3)   \\
\midrule
\textbf{SCAN}       & 49.8±0.5 (51.7) & 80.0±13.4 (95.9) & 48.7±1.1 (49.9)  & 78.0±11.9 (94.0) & 50.2±0.1 (50.3) & 86.8±21.1 (\underline{99.7}) \\
\textbf{Radar}      & 50.2±0.6 (51.0) & 78.4±3.4 (81.6)  & 84.9±3.7 (88.2)  & 59.0±1.7 (61.3)  & OOM\_G          & OOM\_G           \\
\textbf{ANOMALOUS}  & 51.1±1.3 (53.5) & 69.3±16.2 (90.8) & 85.4±0.9 (87.2)  & 59.5±2.5 (62.7)  & OOM\_G          & OOM\_G           \\
\midrule
\textbf{GCNAE}      & \textbf{88.9}±0.0 (\underline{88.9}) & 52.5±0.0 (52.5)  & \textbf{98.6}±0.0 (\underline{98.6})  & 49.0±0.0 (49.0)  & 88.7±9.2 (\underline{94.5}) & 50.0±0.2 (50.3)  \\
\textbf{DOMINANT}   & 71.9±6.6 (74.4) & 93.0±4.6 (95.3)  & 69.0±3.6 (71.3)  & 93.6±2.8 (94.4)  & 69.0±4.5 (71.0) & \textbf{96.3}±10.4 (98.9) \\
\textbf{DONE}       & 70.2±8.3 (80.0) & 92.8±5.6 (\underline{99.8})  & 82.4±11.1 (95.1) & 90.2±8.0 (\underline{99.4})  & 85.7±2.1 (88.7) & 85.5±3.1 (89.1)  \\
\textbf{AdONE}      & 73.9±5.0 (78.0) & 91.3±4.8 (97.3)  & 78.0±10.6 (95.1) & 85.7±11.8 (97.0) & 80.2±4.3 (88.1) & 87.1±1.9 (89.7)  \\
\textbf{AnomalyDAE} & 80.2±2.8 (87.2) & 90.2±4.6 (95.9)  & 88.2±9.6 (98.3)  & 85.5±10.1 (94.3) & 80.0±7.4 (93.3) & 56.6±1.7 (59.0)  \\
\textbf{GAAN}       & 88.7±0.1 (88.8) & 61.0±0.8 (62.5)  & 98.5±0.1 (\underline{98.6})  & 64.2±2.0 (67.3)  & 94.1±0.3 (94.4) & 50.3±0.3 (50.9)  \\
\textbf{GUIDE}      & 88.3±0.8 (88.7) & 61.8±2.4 (71.1)  & OOM\_C           & OOM\_C           & OOM\_C          & OOM\_C           \\
\textbf{CONAD}      & 72.5±5.8 (74.4) & \textbf{93.6}±4.8 (95.4)  & 69.4±2.8 (71.3)  & \textbf{94.1}±0.4 (94.3)  & 65.8±0.9 (67.4) & 68.3±0.6 (69.1) \\
\bottomrule
\end{tabular}}
\vspace{-1.4em}
\end{table}

We report the ROC-AUC metric of different algorithms on three datasets with two types of injected synthetic outliers (contextual and structural outliers from \S \ref{subsec:datasets}) in Table \ref{tab:outlier-types}. Note that we only consider synthetic outliers here since we have generated them so that we know exactly which nodes are contextual vs. structural outliers. Our main findings are as follows. 

\textbf{For GNNs, the reconstruction of the structural information
appears to play a significant role in detecting structural outliers.} Specifically, the performance gap on structural outliers between GCNAE and DOMINANT is over 40\%.
By taking a closer look into these two algorithms, they differ only in that DOMINANT has a structural decoder that aims to reconstruct the adjacency matrix of the graph. That DOMINANT performs significantly better than GCNAE suggests that reconstructing information on graph structure is helpful in identifying structural outliers, which intuitively makes sense as these outliers are defined in terms of graph structure.


\textbf{Low-order structural information (i.e., one-hop neighbors) is sufficient for detecting structural outliers}.
DOMINANT and DONE achieve similar mean ROC-AUC scores ($\sim$92\%) on detecting structural outliers in the Cora and Amazon datasets even though DOMINANT encodes 4-hop neighbor information whereas DONE only encodes 1-hop neighbor information. 
This observation could facilitate outlier node detection model design since encoding high-order information usually imposes a higher computational cost, and multi-hop neighbor aggregation may even lead to the over-smoothing problem in GNNs~\cite{chen2020measuring}.

\textbf{No method achieves high detection accuracy for both structural and contextual outliers.} For instance, none of the methods reaches 85\% detection AUC on both structural and contextual outliers.
\rv{Moreover, on Flickr with structural outliers injected, most attributed graph OD methods that are supposed to detect structural outliers have worse average ROC-AUC scores than that of SCAN, whereas SCAN is a non-attributed graph OD method detecting structural outliers by clustering on nodes.}
The above result suggests that the common approach that arbitrarily combines the structural and contextual loss terms with fixed weights (that are hyperparameters) can struggle to balance performance in detecting both outlier types.
How to detect these two different types of outliers consistently well remains an open question.




We visualize the efficiency in time (wall clock running time) and space (GPU memory consumption) of selected algorithms in Figure~\ref{fig:time_mem}.
The complete results are available in Appx.~\ref{appedix:time_mem}.
All algorithms are evaluated under randomly generated graphs with the same sets of injected outliers to guarantee fairness.
The running time in Figure~\ref{fig:time_mem} (left) is the sum of training and computing outlier scores.
We only measure the GPU memory consumption of the different methods (as opposed to the CPU memory consumption) because it is often the bottleneck of OD algorithms~\cite{zhao2021tod}.
For more details on the generated graphs and experimental settings of this section, see Appx.~\ref{appdx-gen-graph} and Appx.~\ref{appedix:exp_setting}, respectively. Our key findings from Figure~\ref{fig:time_mem} are as follows.

\vspace{-0.125in}
\subsection{Experimental Results: Computational Efficiency}
\label{subsub:efficiency}
\vspace{-0.1in}

\begin{figure}[!t]
\centering
\includegraphics[width=.9\textwidth]{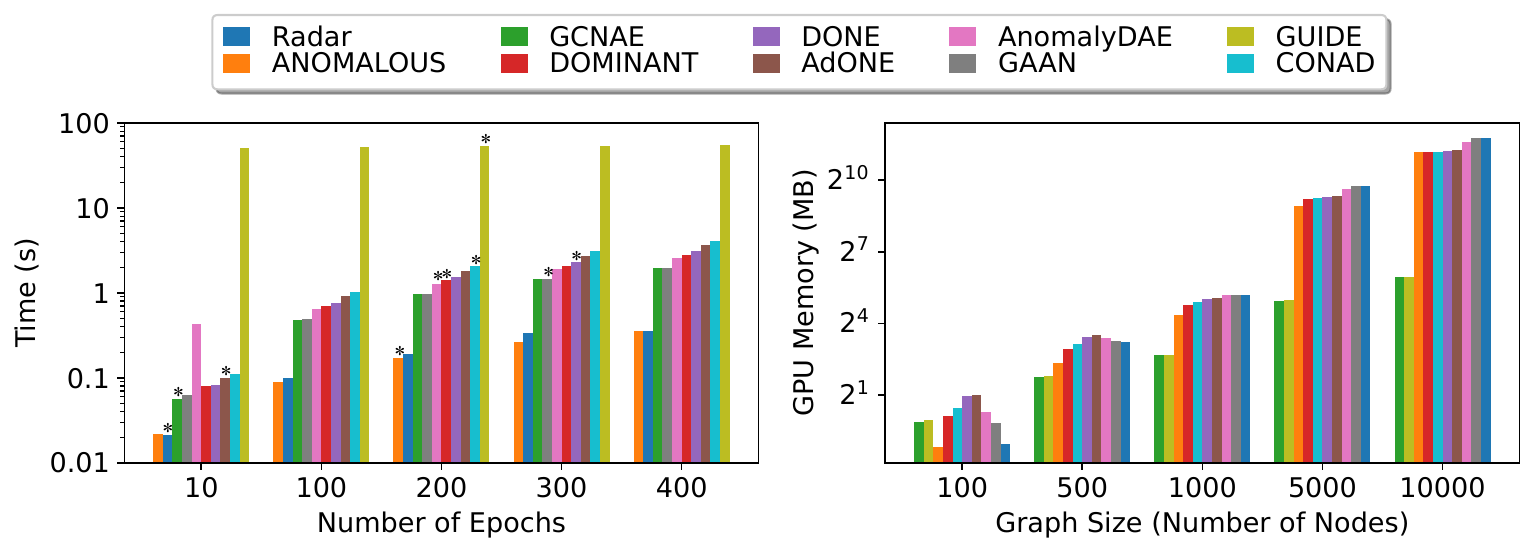}
    \vspace{-.8em}
     \caption{Wall-clock running time (left) and GPU memory consumption (right) of different methods. ($\ast$ denotes the best performance of each method among five different numbers of epochs.)}
     \label{fig:time_mem}
     \vspace{-2em}
\end{figure}

\textbf{Time efficiency.}
Classical methods that we evaluated take less time than the deep learning ones, which tend to be more complicated and learn more flexible models. Among the GNN-based methods, GUIDE consumes far more time compared to others.
The reason is that GUIDE uses a graph motif counting algorithm (which is \#P-complete~\cite{cordella2004sub}) to extract the structural features and consumes much more time on the CPU.
CONAD takes the second-most amount of time due to its use of contrastive learning (which uses pairwise comparisons within mini-batches).

\textbf{Space efficiency.}
According to Figure~\ref{fig:time_mem} (right), GCNAE and GUIDE consume much less GPU memory than the other methods as the graph size increases.
GCNAE saves more memory due to its simpler architecture.
GUIDE consumes more CPU time and RAM to extract low-dimensional node motif degrees, thereby saving more GPU memory.
Though classical methods have the advantage in terms of running time, most of them cannot be deployed in a distributed fashion due to the limitation of ``global'' operators like matrix factorization and inversion.
One advantage of deep models is that they can be easily extended to minibatch and distributed training via graph sampling.
Another advantage is that deep methods can be easily integrated with existing deep learning pipelines (e.g., graph pretraining module that obtains node embeddings).
    

\vspace{-0.1in}

\section{\rv{Discussion}}
\label{sec06:con}
\vspace{-0.1in}

We have established \method, the first comprehensive benchmark for unsupervised outlier node detection on static attributed graphs. 
Our benchmark has empirically examined the effectiveness of a diverse collection of OD algorithms in terms of synthetic vs. organic outliers, structural vs. contextual outliers, and computational efficiency. Importantly, a major goal in our development of \method is to make it easy to extend so that further progress can be made in better understanding existing algorithms and developing new ones to address \prob. We conclude this paper by discussing future research directions for \prob in general (\S \ref{sub:future-opportunities-based-on-experiments}), and then specific to benchmarking (\S \ref{sub:future-work-regarding-benchmarking}).



\vspace{-0.1in}
\subsection{Future Directions in Addressing the Problem \prob}
\label{sub:future-opportunities-based-on-experiments}
\vspace{-0.1in}

Our experimental results on real data (\S \ref{subsub:detection-perform} and \S \ref{subsub:detection-perform}) reveal substantial detection performance differences between algorithms, with none of them being the universal winner. Even for a single algorithm, there is also the issue of hyperparameter tuning (e.g., the detection performance of AnomalyDAE on Weibo varies by as much as 14\% across hyperparameters). However, the fundamental problem is that because the problem \prob is unsupervised, it is not straightforward deciding on the ``right'' choice of algorithm or hyperparameter setting. Any quantitative metric we define to help with model selection or hyperparameter tuning will require assumption(s). Separately, the available computational budget and the size of the dataset to be analyzed can limit what methods can even be used (our results in \S \ref{subsub:efficiency} show that some methods are much more computationally expensive than others).


\rv{\textit{\textbf{Opportunity 1: 
designing ``type-aware'' detection algorithms}}.
A major finding of our experimental results is that which OD algorithm works best heavily depends on the type of outlier encountered (synthetic vs. organic, structural vs. contextual). Put another way, if we expect to see a particular type of outlier, then this should inform the choice of OD algorithm to apply (e.g., MLPAE and GCNAE for contextual outliers). Of course, this would require us to have some a priori knowledge or guess as to what the outliers look like in a dataset.
Importantly, in real applications, we might not need to detect \emph{all} outliers. For example, practitioners may want to 
focus only on high-value or high-interest outliers (e.g., illegal trades that affect revenue the most~\cite{kim2020date} can be considered as contextual outliers, so MLPAE and GCNAE could be good choices). 
Following recent advances in categorizing outlier types in tabular data~\cite{jerez2022equivalence}, we call for attention to identifying more fine-grain outlier types in \prob and figuring out which algorithms are well-suited to these different outlier types. Once we have this sort of information, there could be opportunities for automatically choosing a single or combining multiple OD algorithms, accounting for outlier types (e.g., using an ensembling approach like \cite{zhao2019lscp}).}

\rv{\textit{\textbf{Opportunity 2: synthesizing more realistic and flexible outlier nodes}}. The organic outliers encountered in our real data experiments (\S \ref{subsub:detection-perform}) can be complex and composed of multiple outlier types (possibly including types beyond just the structural and contextual ones we focused on). Our experimental results show that there is a wide detection performance gap on synthetic vs. organic outliers, calling for more realistic outlier generation approaches. To improve existing generation methods in \method, we could use a generative model to fit the normal samples, and then perturb the generative model to generate different types of outliers. This approach has been successful in tabular OD \cite{steinbuss2021benchmarking}.}

\rv{\textit{\textbf{Opportunity 3: understanding the sensitivity of \prob algorithms to hyperparameters and designing more stable methods}}.
We had pointed out in \S \ref{subsub:detection-perform} that some deep learning methods are more stable (across hyperparameters) than others in terms of standard deviation in achieved ROC-AUC scores. This phenomenon extends of course to classical methods as well.
Better understanding the drivers of these algorithms' (in)sensitivity to hyperparameters would help us better design algorithms that are more stable with respect to hyperparameter settings. In turn, this could help ease the burden of unsupervised hyperparameter tuning.
In tabular OD tasks, researchers have developed methods for outlier detection that are more stable with respect to hyperparameters such as the robust autoencoder \cite{zhou2017anomaly}, RandNet \cite{chen2017outlier}, ROBOD \cite{ding2022hyperparameter}, and ensemble frameworks \cite{zhao2019lscp}. Perhaps some techniques from these tabular OD methods could be incorporated into specific methods for \prob to improve stability.}

\rv{\textit{\textbf{Opportunity 4: 
developing more efficient \prob algorithms.}}
Our results on computational efficiency (\S \ref{subsub:efficiency}) show that some algorithms take substantially more time and/or memory to execute than others.
Meanwhile, most algorithms tested ran out of memory on the million-scale DGraph dataset (Table~\ref{tab:main-table-real}).
We suggest developing more scalable algorithms for \prob, which could mean more optimized implementations of existing algorithms and also the development of new algorithms. We point out several lines of work that could be helpful in this endeavor.
First, there are existing approaches for making GNNs more scalable \cite{jia2020improving,li2021training} but these have yet to be specialized to address \prob.
Separately, most existing autoencoder-based methods that we tested reconstruct a complete graph adjacency matrix (e.g., DOMINANT, DONE, AnomalyDAE, GAAN), which is memory intensive (scaling quadratically with the number of nodes). Developing a more memory-efficient implementation of this step would be interesting.
Next, approximating the node motif degree in GUIDE is possible, which can significantly reduce both computation time and space \cite{chen2020can, yu2021count}.
Lastly, we mention that some recent work \cite{zhao2021suod,zhao2021tod} accelerates tabular OD via distributed learning, data and model compression, and/or quantization. These ideas could be extended to algorithms for \prob.
}

\rv{\textbf{\textit{Opportunity 5: meta-learning to assist model selection and hyperparameter tuning}}.
Recent work on general graph learning \cite{park2022autogml} and unsupervised OD model selection on tabular data \cite{zhao2021automatic} shows that under meta-learning frameworks, we can identify good OD models to use for a new task (or dataset) based on its similarity to meta-tasks where ground truth information is available. A similar approach also works for unsupervised hyperparameter tuning \cite{zhao2022towards}.
We suggest
exploring a meta-learning framework for algorithm selection and hyperparameter tuning in solving \prob, including quantifying task similarity between graph OD datasets. Such a framework would require some but not all datasets to have ground truth, and that datasets with ground truth can be related to the ones without.}

\vspace{-0.1in}
\subsection{Future Directions in Improving Our Benchmark System \method}
\label{sub:future-work-regarding-benchmarking}
\vspace{-0.1in}

\textbf{Extending detection tasks to different ``levels''.} In \method, we focus on node-level detection with \rv{static attributed} graphs due to their popularity, while there are more detection tasks at different levels of a graph. Recent graph OD algorithms extend to edge- \cite{zhang2022efraudcom}, subgraph- \cite{wang2018deep}, and graph-level~\cite{qiu2022Raising,zhao2021using} detection. 
\rv{Future comprehensive graph OD benchmarks can include these emerging graph OD tasks.}

\textbf{Incorporating supervision.}
Although \method focuses on unsupervised methods,
there can be cases where a small set of labels (either for OD or relevant tasks) are available so that (semi-)supervised learning is possible (e.g., \cite{dou2020enhancing,zhang2021fraudre}). Extending \method to handle supervision would particularly be beneficial in addressing algorithm selection and hyperparameter tuning challenges.

\textbf{Curating more datasets.}
\rv{Thus far, we have only included nine real datasets in \method.} Adding more datasets over time would be beneficial, especially ones with organic outliers.
With a much larger collection of datasets (e.g., $\ge20$), one could run 
statistical tests for comparison \cite{demvsar2006statistical}, which has been used in OD tasks with tabular datasets \cite{li2022ecod,zhao2021automatic}.
Similar to tabular OD \cite{emmott2015meta}, one can convert existing multi-classification graph datasets (e.g., ones from Open Graph Benchmark (OGB) \cite{ogb}) into OD datasets by treating one or combining several small classes to be treated as a single ``outlier'' class, with all other classes considered ``normal''.

\section*{Acknowledgments}
\rv{\textbf{Funding.} This work was supported in part by NSF under grants III-1763325, III-1909323, III-2106758, and SaTC-1930941. K.S. is supported by a Cisco Research Award. G.H.C.~is supported by NSF CAREER award \#2047981.

\textbf{Author contributions.}
Conceptualization: K.L., Y.D., and Y.Z. Investigation and experiments: K.L., Y.D., Y.Z., X.D., X.H., R.Z., K.D., and C.C.
Writing – original draft: K.L., Y.D., and Y.Z.
Writing – review \& editing: H.P., K.S., L.S., J.L., Z.H., and P.S.Y.
Extensive reviewing \& editing: G.H.C.}

For any correspondence, please refer to Hao Peng.

\bibliographystyle{abbrv}
\small
\bibliography{sample}

\clearpage
\newpage

\appendix
\normalsize
\section{Additional Details on \method}

\subsection{Additional Dataset Information}
\label{appdx-dataset}

\subsubsection{Real Data}
\textbf{Cora}~\cite{sen2008collective} is a citation graph with nodes representing machine learning papers and edges representing papers' citation relationships.
The node features are sparse bag-of-words (BoW) vectors extracted from the paper document, and their labels represent one of the seven classes.

\textbf{Amazon}~\cite{shchur2018pitfalls} is a segment of the Amazon co-purchase graph~\cite{mcauley2015image}, where nodes represent goods and edges indicate that two goods are frequently bought together. Notably, node features are BoW-encoded product reviews, and class labels are given by the product category.

\textbf{Flickr}~\cite{zeng2019graphsaint} datasets 
originates from NUS-wide\cite{chua2009nus}, a real-world web image database
from the National University of Singapore. 
The SNAP website\footnote{\url{http://snap.stanford.edu/}} collected Flickr data from four different sources including NUS-wide, and generated an undirected graph.
A node in the graph represents one image uploaded to Flickr. 
If two images share some common properties (e.g., same geographic location, same gallery, comments by the same user, etc.), one edge is made between these two nodes.
The node feature is composed of a 500-dimensional 
vector of the images provided by NUS-wide.
Eighty-one tags of each image are manually merged into seven classes, and each image is assigned to one of the seven classes.

\textbf{Weibo}~\cite{zhao2020error} is a user-posts-hashtag graph from Tencent-Weibo, a Twitter-like platform in China.
This dataset collects information from 8,405 users with 61,964 hashtags. We use the user-user graph~\footnote{\url{https://github.com/zhao-tong/Graph-Anomaly-Loss/tree/master/data/weibo_s}} provided by the author, which connects users who used the same hashtag.
Temporal information was used to label the users.
If a user made at least five suspicious events, he/she is labeled as a suspicious user; if no suspicious event was made, he/she is a benign user.
There are a total of 868 suspicious users and 7,537 benign users.
The suspicious users are regarded as outliers in the graph.
Since the ground truth was generated using time information, the timestamps are not used to create raw user features.
Therefore, the raw feature vector has two parts:
(1) for each user, the one-hot vectors of his/her posts are summed where each one-hot vector represents the location where a post was made. Then the \#dimension of the summed vector is reduced to 100 using SVD and
(2) for each user, the \#dimension of the BoW vectors extracted from post texts is reduced to 300.
The final node feature is the concatenation of the location vector and the BoW vector.
Note that Weibo is a directed graph; the remaining datasets used in our benchmark are undirected graphs.

\textbf{Reddit}~\cite{kumar2019predicting, wang2021decoupling} is a user-subreddit graph extracted from a social media platform, Reddit~\footnote{\url{https://www.reddit.com/}}.
This public dataset consists of one month
of user posts on subreddits~\footnote{\url{http://files.pushshift.io/reddit/}}. 
The 1,000 most active subreddits and the 10,000 most active users are extracted as subreddit nodes and user nodes, respectively.
This results in 168,016 interactions.
Each user has a binary label indicating whether it has been banned by the platform.
We assume that the banned users are outliers compared to normal Reddit users.
The text of each post is converted into a feature vector representing their LIWC categories~\cite{pennebaker2001linguistic} and the features of users and subreddits are the feature summation of the posts they have, respectively.

\rv{\textbf{Disney}~\cite{muller2013ranking} and \textbf{Books}~\cite{sanchez2013statistical} come from the Amazon co-purchase networks~\cite{leskovec2007dynamics}.
Disney is a co-purchase network of movies, where the attributes include prices, ratings, number of reviews, etc. The ground truth labels (i.e., whether it is an outlier) are manually labeled by high school students by majority vote.
The second dataset, Books, is a co-purchase network of books on Amazon, which has similar attributes to the Disney dataset.
The ground truth labels are derived from 
$\texttt{amazonfail}$ tag information.
More information about the datasets can be found on the project website~\footnote{\url{https://www.ipd.kit.edu/~muellere/consub/}}.
}

\rv{\textbf{Enron}~\cite{sanchez2013statistical} is an email network dataset extracted from~\cite{klimt2004enron}. Each email is regarded as a node, and the messages between email addresses represent edges. The email addresses having sent spam messages are taken as outliers. Each node contains 20 attributes describing aggregated information about the average content length, the average number of recipients, or the time range between two emails.}

\rv{\textbf{DGraph}~\cite{huang2022dgraph} is a large-scale attributed graph with 3M nodes, 4M dynamic edges, and 1M ground-truth nodes. The nodes represent user accounts in a financial company providing personal loan services, and the edge between two nodes represents one account that has added another account as an emergency contact. For all the accounts with at least one borrowing record, the outliers are the accounts with overdue history, and the inliers are the accounts without overdue. 
Note there are also 2M accounts/nodes without any borrowing at all.
The 17 node features are encoded from the user profile information like age and gender.}

\subsubsection{Random Graph Generation Method}
\label{appdx-gen-graph}

We leverage a random graph generation method used in~\cite{kim2022find} to create an arbitrary \prob graph for benchmarking.
Specifically, the implementation in PyG~\footnote{\url{https://pytorch-geometric.readthedocs.io/en/latest/modules/datasets.html\#torch\_geometric.datasets.RandomPartitionGraphDataset}} is used with 2 classes, node\_homophily\_ratio=0.5, average\_degree=5 and num\_channels=64.
We use the generated random graphs to benchmark algorithms' efficiency and scalability.
We generate a random graph \textbf{Gen\_Time} with num\_nodes\_per\_class=500 (1000 in total) as the graph data to test the runtime.
To benchmark the scalability, we generate multiple random graphs \textbf{Gen\_100}, \textbf{Gen\_500}, \textbf{Gen\_1000}, \textbf{Gen\_5000} and \textbf{Gen\_10000} with num\_nodes\_per\_class equal to 50, 250, 500, 2500 and 5000, respectively.

\subsubsection{Outlier Injection Details}
\label{outlier-inject}



For injecting structural outliers, we use $p=0.2$. For contextual outliers, we set $q$ equal to $m$ in structural outlier injection.
The other parameters used in outlier injection are shown in Table~\ref{tab:inject}. Note that $m$ is set to be approximate twice the degree of the graph. For real datasets, we keep a similar outlier ratio (i.e., the number of outliers injected is approximately 5\% of the total number of nodes; as a reminder, for structural outliers, we inject $m\times n$ outliers and for contextual outliers, we inject $o$ outliers; $o=m\times n$ in our setting). We keep a similar number of outliers for generated graph datasets of various sizes. The statistics of the generated graphs are shown in Table~\ref{tab:gen-stat}.

\begin{table}[h]
\centering
\caption{Parameters used in synthetic outliers injection.}
\label{tab:inject}
\scalebox{0.8}{
\begin{tabular}{@{}cccccccccc@{}}
\toprule
                & \textbf{Cora} & \textbf{Amazon} & \textbf{Flickr} & \textbf{Gen\_Time} & \textbf{Gen\_100} & \textbf{Gen\_500} & \textbf{Gen\_1000} & \textbf{Gen\_5000} & \textbf{Gen\_10000} \\ \midrule
\textbf{Degree} & 4.1           & 37.5            & 10.6            & 5            & 5                 & 5                 & 5                  & 5                  & 5                   \\
\textbf{n}      & 70            & 350             & 2240            & 10           & 1                 & 1                 & 1                  & 1                  & 1                   \\
\textbf{m}      & 10            & 70              & 20              & 10           & 10                & 10                & 10                 & 10                 & 10                  \\ \bottomrule
\end{tabular}}
\end{table}
\begin{table}[!ht]
\centering
\caption{Statistics of generated datasets in \method.}
\label{tab:gen-stat}
\scalebox{0.92}{
\begin{tabular}{@{}l|cccccccc@{}}
\toprule
\textbf{Dataset} & \textbf{\#Nodes} & \textbf{\#Edges} & \textbf{\#Feat.   } & \textbf{Degree} & \textbf{\#Con.} & \textbf{\#Strct.} & \textbf{\#Outliers} & \textbf{Ratio}  \\ \midrule
\textbf{Gen\_Time}   & 1,000            & 5,746           & 64                  & 5.7             & 100              & 100                       & 189            & 18.9\%            \\
\textbf{Gen\_100}    & 100           & 618          & 64                 & 6.2            & 10             & 10                      & 18            & 18.0\%            \\
\textbf{Gen\_500}    & 500           & 2,662          & 64                 & 5.3            & 10           & 10                    & 20          & 4.0\%            \\
\textbf{Gen\_1000}   & 1,000            & 4,936          & 64                 & 4.9            & 10               & 10                        & 20            & 2.0\%           \\
\textbf{Gen\_5000}   & 5,000           & 24,938          & 64                  & 5.0            & 10               & 10                        & 20            & 0.4\%            \\
\textbf{Gen\_10000}  & 10,000           & 49,614          & 64                  & 5.0            & 10               & 10                        & 20            & 0.2\%            \\
\bottomrule
\end{tabular}}
\vspace{-0.1in}
\end{table}

\subsection{Description of algorithms in the benchmark}
\label{appdx-algorithms}

\textbf{LOF~\cite{breunig2000lof}.} LOF is short for the Local Outlier Factor. LOF computes the degree of an object as abnormality, and the degree depends on how isolated the object is with respect to its surrounding neighborhood.
Note that LOF only uses node attribute information, and the neighborhood is composed of $k$-nearest-neighbors.

\textbf{IF~\cite{liu2012isolation}.} Isolation Forest (IF) is a classic tree ensemble method used in outlier detection.
It builds an ensemble of base trees to isolate the data points and defines the decision boundary as the closeness of an individual instance to the root of the tree.
It only uses node attributes of data.

\textbf{MLPAE~\cite{sakurada2014anomaly}.}
The MLPAE is a vanilla autoencoder with multiple layer perceptron (MLP) as encoder and decoder. The encoder takes the node attribute as the input to learn its low-dimensional embedding and a decoder reconstructs the input node attribute from the node embedding.
The outlier score of a node is the reconstruction error of the decoder.

\textbf{SCAN~\cite{xu2007scan}.}
SCAN is a structural clustering algorithm to detect clusters, hub nodes, and outlier nodes in a graph.
Since the structural outliers exhibit clustering patterns on graphs, we use SCAN to detect clusters and the nodes in detected clusters are regarded as structural outliers in the graph.
SCAN only takes the graph structure as the input.

\textbf{Radar~\cite{li2017radar}.}
Radar is an anomaly detection framework for attributed graphs.
It takes the graph structure and node attributes as the input.
It detects outlier nodes whose behaviors are singularly different from the majority by characterizing the residuals of attribute information and its coherence with network information.
The outlier score of a node is decided by the norm of its reconstruction residual.

\textbf{ANOMALOUS~\cite{peng2018anomalous}.}
ANOMALOUS performs joint anomaly detection and attribute selection to detect node anomalies
on attributed graphs based on the CUR decomposition and residual analysis.
It takes the graph structure and node attribute as the input, and the outlier score of a node is decided by the norm of its reconstruction residual.

\textbf{GCNAE~\cite{kipf2016variational}.}
GCNAE is the autoencoder framework with GCNs~\cite{kipf2016semi} as the encoder and decoder.
It takes the graph structure and node attributes as input.
The encoder is used to learn a node's embedding by aggregating its neighbor information.
The decoder reconstructs the node attribute by applying another GCN to node embeddings and graph structures.
Similar to MLPAE, the outlier score of a node is the reconstruction error of the decoder.

\textbf{DOMINANT~\cite{ding2019deep}.}
DOMINANT is one of the first works that leverage GCN and AE for outlier node detection. It uses a two-layer GCN as the encoder, a two-layer GCN decoder to reconstruct the node attribute, and a one-layer GCN and dot product as the structural decoder to reconstruct the graph adjacency matrix.
The reconstruction errors of both decoders are combined as the outlier scores of the nodes.

\textbf{DONE~\cite{bandyopadhyay2020outlier}.}
DONE leverages a structural and an attribute AE to reconstruct the adjacency matrix and node attribute. The encoders and decoders are composed of MLPs.
The node embeddings and outlier scores are optimized simultaneously with a unified loss function.

\textbf{AdONE~\cite{bandyopadhyay2020outlier}.}
AdONE is a variant of DONE, which uses an extra discriminator to discriminate the learned structure embedding and attribute embedding of a node.
The adversarial training approach supposes to better align the two different embeddings in the latent space.

\textbf{AnomalyDAE~\cite{fan2020anomalydae}.}
AnomalyDAE also utilizes a structure AE and attribute AE to detect outlier nodes.
The structure encoder of AnomalyDAE takes both the adjacency matrix and node attribute as input; the attribute decoder reconstructs the node attribute using both structure and attribute embeddings.

\textbf{GAAN~\cite{chen2020generative}.}
GAAN is a GAN-based outlier node detection method. It employs an MLP-based generator to generate fake graphs and an MLP-based encoder to encode graph information.
A discriminator is trained to recognize whether two connected nodes are from the real or fake graph.
The outlier score is obtained by the node reconstruction error and real-node identification confidence.

\textbf{GUIDE~\cite{yuan2021higher}.}
GUIDE is similar to DONE and AdONE with two different AEs, but it pre-processes the structure information before feeding it into the structure encoder.
Specifically, node motif degree is used to represent the node structure vector which could encode higher-order structure information.

\textbf{CONAD~\cite{xu2022contrastive}.}
CONAD is one of the \method methods that leverage graph augmentation and contrastive learning techniques.
It imposes prior knowledge of outlier nodes via generating augmented graphs.
After encoding the graphs using Siamese GNN encoders, the contrastive loss is used to optimize the encoder, and the outlier score of the node is obtained by two different decoders like DOMINANT.

\subsection{Description of Evaluation Metrics}
\label{appdx-metrics}

\textbf{ROC-AUC (AUC).}
AUC computes the Area Under the Receiver Operating Characteristic Curve (ROC-AUC) from predicted outlier scores.
The ROC curve is created by plotting the true positive rate (TPR) against the false positive rate (FPR) at various threshold settings.
In the \method benchmark, we regard the outlier nodes as the positive class and compute the AUC for it.
AUC equals 1 means the model makes a perfect prediction, and AUC equals 0.5 means the model has no class-separation capability.
AUC is better than accuracy when evaluating the outlier detection task since it is not sensitive to the imbalanced class distribution of the data.

\textbf{Average Precision (AP).}
AP summarizes the precision-recall curve as the weighted mean of precisions achieved at each threshold, with the increase in recall from the previous threshold used as the weight.
AP is a metric that balances the effects of recall and precision, and a higher AP indicates a lower false-positive rate (FPR) and false-negative rate (FNR).
FPR and FNR have equal importance for most outlier detection applications as more misclassified normal samples could worsen legit users' experience. 

\textbf{Recall@k.}
The outliers are usually rare in contrast to enormous normal samples in the data, and the outliers are of the most interest to outlier detection practitioners.
We propose to use Recall@k to measure how well the detectors rank outliers over the normal samples.
We set $k$ as the number of ground truth outliers in each dataset.
The Recall@k is computed by the number of true outliers among the top-k samples in the outlier ranking list divided by $k$.
A higher Recall@k score indicates a better detection performance, and Recall@k equals 1 means the model perfectly ranks all outliers over the normal samples.

\textbf{Runtime.} Due to the coverage of both classical algorithms and neural network methods, we consistently measure the model runtime as the duration between the experiment starts and ends, mimicking the real-world applications without specific differentiation between CPU and GPU time.

\textbf{GPU Memory.} Notably, GPU memory is often the bottleneck of machine learning algorithms due to its limitation in extension. In \method, we report the max active GPU memory for running an algorithm.


\section{Additional Experimental Settings and Details}
\label{appedix:exp_setting}

\textbf{Environment.}
The key libraries and their versions used in the experiment are as follows: Python=3.7, CUDA\_version=11.1, torch=1.10, pytorch\_geometric>=2.0.3, networkx=2.6.3, numpy=1.19.4, scipy=1.5.2, scikit-learn=0.22.1, pyod=1.0.1, pygod=0.3.0.

\textbf{Hardware configuration.}
All the experiments were performed on a Linux server with a 3.50GHz Intel Core i5 CPU, 64GB RAM, and 1 NVIDIA GTX 1080 Ti GPU with 12GB memory.

\textbf{More model implementation details.}
To include a large number of algorithms, we build Python Graph Outlier Detection (PyGOD)\footnote{PyGOD: \url{https://pygod.org/}} \cite{liu2022pygod}, which provides more than 10 latest graph OD algorithms; all with unified APIs and optimizations.
We tried our best to apply the same set of optimization techniques to each dataset. For Radar and ANOMALOUS, we use gradient descent instead closed-form optimization provided in official implementation due to fairness and efficiency concerns. For all deep algorithms, we implement sampling and minibatch training on large graphs (e.g., Flickr). See our library source code for more details.
Meanwhile, we also include multiple non-graph baselines (LOF and IF) from our early work Python Outlier Detection (PyOD) \cite{zhao2019pyod}. 

\textbf{Hyperparameter grid}

The hyperparameter space is shown in Table~\ref{tab:param}. The candidates of hyperparameters are listed in square brackets. In each trial, a value is randomly chosen among candidates. The results (mean, std, max) are reported among 20 trials.

Due to the large graph size, full batch training on Flickr cannot fit in single GPU memory. Minibatch training and different batch size, sampling size, and the number of epochs are used on Flickr. 
Because of the complexity of real datasets, automated balancing by the standard deviation for weight alpha cannot balance well. Thus, three candidates are attempted.
As Reddit has a lower feature dimension, we reduce hidden dimension values on Reddit.

\begin{table}[!ht]
\centering
\caption{Hyperparameters in different algorithms. The values in "[]" are candidates. We present these common hyperparameters shared by multiple algorithms on the top, and also specify some algorithm-specific hyperparameters at the bottom. Refer to PyGOD doc for more details.}
\label{tab:param}
\scalebox{0.7}{
\begin{tabular}{|c|c|cccccccccc|}
\hline
\textbf{Algorithm}                   & \textbf{Hyperparameter} & \multicolumn{1}{c|}{\textbf{Cora}} & \multicolumn{1}{c|}{\textbf{Amazon}} & \multicolumn{1}{c|}{\textbf{Flickr}} & \multicolumn{1}{c|}{\textbf{Weibo}} & \multicolumn{1}{c|}{\textbf{Disney}} & \multicolumn{1}{c|}{\textbf{Books}}& \multicolumn{1}{c|}{\textbf{Enron}} & \multicolumn{1}{c|}{\textbf{DGraph}} & \multicolumn{1}{c|}{\textbf{Reddit}} & \textbf{Gen} \\ \hline
\multirow{8}{*}{\textbf{Common}}     & \textit{dropout}        & \multicolumn{10}{c|}{{[}0, 0.1, 0.3{]}}                                                                                                                                                                       \\ \cline{2-12} 
                                     & \textit{learning rate}  & \multicolumn{10}{c|}{{[}0.1, 0.05, 0.01{]}}                                                                                                                                                                   \\ \cline{2-12} 
                                     & \textit{weight decay}   & \multicolumn{10}{c|}{0.01}                                                                                                                                                                                    \\ \cline{2-12} 
                                     & \textit{batch size}     & \multicolumn{2}{c|}{full batch}                                           & \multicolumn{1}{c|}{64}              & \multicolumn{4}{c|}{full batch}   & \multicolumn{1}{c|}{64}       & \multicolumn{2}{c|}{full batch}                                 \\ \cline{2-12} 
                                     & \textit{sampling}       & \multicolumn{2}{c|}{all neigh.}                                           & \multicolumn{1}{c|}{3}               & \multicolumn{4}{c|}{all neigh.}   & \multicolumn{1}{c|}{3}               & \multicolumn{2}{c|}{all neigh.}                               \\ \cline{2-12} 
                                     & \textit{epoch}          & \multicolumn{2}{c|}{300}                                                  & \multicolumn{1}{c|}{2}               & \multicolumn{4}{c|}{300}                                                  & \multicolumn{1}{c|}{2}  & \multicolumn{2}{c|}{300}            \\ \cline{2-12} 
                                     & \textit{alpha}          & \multicolumn{3}{c|}{auto}                                                                                        & \multicolumn{6}{c|}{{[}0.8, 0.5, 0.2{]}}                                   & auto         \\ \cline{2-12} 
                                     & \textit{hid. dim.}      & \multicolumn{4}{c|}{{[}32, 64, 128, 256{]}} & \multicolumn{4}{c|}{{[}8, 12, 16{]}}                                                       & \multicolumn{2}{c|}{{[}32, 48, 64{]}}               \\ 
                                     \hline\hline
\multirow{2}{*}{\textbf{SCAN}}       & \textit{eps}            & \multicolumn{10}{c|}{{[}0.3, 0.5, 0.8{]}}                                                                                                                                                                     \\ \cline{2-12} 
                                     & \textit{mu}             & \multicolumn{10}{c|}{{[}2, 5, 10{]}}                                                                                                                                                                          \\ \hline
\multirow{2}{*}{\textbf{AnomalyDAE}} & \textit{theta}          & \multicolumn{10}{c|}{{[}10, 40, 90{]}}                                                                                                                                                                        \\ \cline{2-12} 
                                     & \textit{eta}            & \multicolumn{10}{c|}{{[}3, 5, 8{]}}                                                                                                                                                                           \\ \hline
\textbf{GAAN}                        & \textit{noise dim.}     & \multicolumn{10}{c|}{{[}8, 16, 32{]}}                                                                                                                                                                         \\ \hline
\textbf{GUIDE}                       & \textit{struct. hid.}   & \multicolumn{10}{c|}{{[}4, 5, 6{]}}                                                                                                                                                                           \\ \hline
\end{tabular}
}
\end{table}

\textbf{How we determine the optimal performance in runtime comparison.}

The optimal performance is determined by the ROC-AUC score. Taking the computational cost into account, we expect a reasonable score within as few training epochs as possible. Thus, when the score converges (i.e., the score increment of consequence epochs is less than 0.5\%), we mark the current epoch as optimal.

\clearpage
\newpage

\section{Additional Experimental Results}
\label{appedix:add_results}

\subsection{Additional Results on Real Dataset Detection Performance}
\vspace*{\fill}
\begin{table}[!ht]
\centering
\caption{\small Average Precision (\%) comparison among OD algorithms on three datasets with synthetic outliers, where we show \textit{the avg perf.} $\pm$ \textit{the STD of perf.} (\textit{max perf.}) of each. The best algorithm by expectation is shown in \textbf{bold}, while the max performance per dataset is marked with \underline{underline}. OOM denotes out of memory with regard to GPU (\_G) and CPU (\_C).}
\label{tab:main-ap}
\scalebox{0.74}{
\begin{tabular}{@{}l|ccc@{}}
\toprule
\textbf{Algorithm} & \textbf{Cora}         & \textbf{Amazon}       & \textbf{Flickr}  \\ \midrule
\textbf{LOF}        & 12.2±0.0 (12.2) & 5.6±0.0 (5.6) & 5.4±0.0 (5.4) \\
\textbf{IF}      & 10.1±0.7 (11.5) & 6.2±1.3 (9.5)  & 8.2±0.6 (9.3)           \\
\textbf{MLPAE}  & 13.2±0.0 (13.2) & \textbf{34.8}±0.0 (34.9)  & 18.7±0.0 (18.7)          \\
\midrule
\textbf{SCAN}       & 12.0±5.8 (21.6) & 13.8±10.5 (33.8) & 24.6±20.6 (\underline{50.4}) \\
\textbf{Radar}      & 7.5±0.4 (7.9)   & 12.3±1.5 (14.2)  & OOM\_G           \\
\textbf{ANOMALOUS}  & 5.9±1.7 (8.8)   & 11.7±1.3 (15.3)  & OOM\_G           \\
\midrule
\textbf{GCNAE}      & 13.2±0.0 (13.2) & \textbf{34.8}±0.0 (34.9)  & 18.0±2.8 (18.6)  \\
\textbf{DOMINANT}   & 20.0±3.0 (20.8) & 16.0±0.9 (16.6)  & \textbf{28.6}±11.8 (36.4) \\
\textbf{DONE}       & \textbf{25.0}±8.8 (\underline{42.4}) & 19.3±7.7 (\underline{36.5})  & 20.0±2.7 (24.9)  \\
\textbf{AdONE}      & 19.3±4.2 (29.2) & 23.7±4.7 (31.2)  & 18.2±3.5 (25.1)  \\
\textbf{AnomalyDAE} & 18.3±2.1 (21.3) & 24.0±7.2 (33.4)  & 12.3±3.8 (18.2)  \\
\textbf{GAAN}       & 14.6±0.4 (15.1) & 34.5±0.3 (34.8)  & 18.6±0.1 (18.7)  \\
\textbf{GUIDE}      & 14.0±0.5 (14.8) & OOM\_C           & OOM\_C           \\
\textbf{CONAD}      & 17.1±5.5 (20.7) & 15.5±2.0 (16.6)  & 8.8±1.1 (10.0)   \\
\bottomrule
\end{tabular}}
\vspace{-0.2in}
\end{table}
\vspace*{\fill}
\begin{table}[!ht]
\centering
\caption{\small Average Precision (\%) comparison among OD algorithms on six datasets with organic outliers, where we show \textit{the avg perf.} $\pm$ \textit{the STD of perf.} (\textit{max perf.}) of each. The best algorithm by expectation is shown in \textbf{bold}, while the max performance per dataset is marked with \underline{underline}. OOM denotes out of memory with regard to GPU (\_G) and CPU (\_C). TLE denotes time limit of 24 hours exceeded.}
\label{tab:main-ap-real}
\scalebox{0.74}{
\begin{tabular}{@{}l|cccccc@{}}
\toprule
\textbf{Algorithm} & \textbf{Weibo}  & \textbf{Reddit}   & \textbf{Disney}   & \textbf{Books}       & \textbf{Enron}     & \textbf{DGraph}      \\ 
\midrule
\textbf{LOF}             & 15.8±0.0 (15.8)  & \textbf{4.2}±0.0 (4.2) & 5.2±0.0 (5.2)   & 1.5±0.0 (1.5) & 0.0±0.0 (0.0) & TLE           \\
\textbf{IF}              & 12.9±2.6 (19.8)  & 2.8±0.1 (2.9) & \textbf{10.1}±4.5 (22.6) & 1.9±0.2 (2.7) & 0.1±0.0 (0.1) & \textbf{1.8}±0.0 (\underline{1.9}) \\
\textbf{MLPAE}           & 52.8±9.9 (64.5)  & 3.4±0.0 (3.4) & 5.9±0.8 (7.9)   & 1.8±0.3 (2.5) & 0.1±0.0 (0.1) & 0.9±0.0 (1.0) \\
\midrule
\textbf{SCAN}            & 17.3±3.4 (20.5)  & 3.3±0.0 (3.3) & 5.0±0.3 (5.5)   & 2.0±0.1 (2.1) & 0.0±0.0 (0.1) & TLE           \\
\textbf{Radar}           & \textbf{92.1}±0.7 (\underline{92.9})  & 3.6±0.2 (3.9) & 7.2±0.0 (7.2)   & 2.2±0.0 (2.2) & \textbf{0.2}±0.0 (0.2) & OOM\_C        \\
\textbf{ANOMALOUS}       & \textbf{92.1}±0.7 (\underline{92.9})  & 4.0±0.6 (\underline{5.1}) & 7.2±0.0 (7.2)   & 2.2±0.0 (2.2) & \textbf{0.2}±0.0 (0.2) & OOM\_C        \\
\midrule
\textbf{GCNAE}           & 70.8±5.0 (80.9)  & 3.4±0.0 (3.4) & 4.8±0.7 (5.8)   & 2.1±0.4 (3.5) & 0.1±0.0 (0.1) & 1.0±0.0 (1.0) \\
\textbf{DOMINANT}        & 18.0±10.2 (36.2) & 3.7±0.0 (3.8) & 7.6±5.0 (\underline{23.2})  & 2.2±0.6 (4.1) & 0.1±0.1 (\underline{0.4}) & OOM\_C        \\
\textbf{DONE}            & 65.5±13.4 (77.3) & 3.7±0.4 (4.5) & 5.0±0.7 (6.4)   & 1.8±0.3 (2.6) & 0.1±0.0 (0.1) & OOM\_C        \\
\textbf{AdONE}           & 62.9±9.5 (74.4)  & 3.3±0.4 (4.0) & 6.1±1.5 (11.7)  & 2.5±0.3 (3.2) & 0.1±0.0 (0.1) & OOM\_C        \\
\textbf{AnomalyDAE}      & 38.5±22.5 (77.3) & 3.7±0.1 (3.8) & 5.7±0.2 (6.3)   & \textbf{3.5}±1.4 (\underline{7.8}) & 0.1±0.0 (0.1) & OOM\_C        \\
\textbf{GAAN}            & 80.3±0.2 (80.7)  & 3.7±0.1 (3.9) & 5.6±0.0 (5.6)   & 2.6±0.8 (5.6) & 0.1±0.0 (0.1) & OOM\_C        \\
\textbf{GUIDE}           & OOM\_C           & OOM\_C        & 4.8±0.9 (6.9)   & 1.9±0.3 (3.1) & OOM\_C        & OOM\_C        \\
\textbf{CONAD}           & 15.6±6.9 (31.7)  & 3.7±0.3 (4.6) & 6.0±1.4 (11.5)  & 2.5±0.8 (4.9) & 0.1±0.0 (0.3) & 0.9±0.0 (0.9) \\
\bottomrule
\end{tabular}}
\end{table}
\vspace*{\fill}

\clearpage
\newpage
\vspace*{\fill}
\begin{table}[!ht]
\centering
\caption{\small Recall@k (\%) comparison among OD algorithms on three datasets with synthetic outliers, where we show \textit{the avg perf.} $\pm$ \textit{the STD of perf.} (\textit{max perf.}) of each. The best algorithm by expectation is shown in \textbf{bold}, while the max performance per dataset is marked with \underline{underline}. k is set as the number of outliers in labels. OOM denotes out of memory with regard to GPU (\_G) and CPU (\_C).}
\label{tab:main-rec}
\scalebox{0.74}{
\begin{tabular}{@{}l|ccc@{}}
\toprule
\textbf{Algorithm} & \textbf{Cora}         & \textbf{Amazon}       & \textbf{Flickr} \\ \midrule
\textbf{LOF}        & 15.2±0.0 (15.2)  & 5.2±0.0 (5.2)    & 8.4±0.0 (8.4)   \\
\textbf{IF}         & 14.6±1.7 (18.1)  & 5.0±2.0 (9.9)    & 12.6±1.2 (14.7) \\
\textbf{MLPAE}       & 18.8±0.0 (18.8)  & 45.0±0.0 (45.1)  & 28.0±0.1 (28.1)\\
\midrule
\textbf{SCAN}       & 17.5±8.9 (32.6) & 16.1±11.2 (32.7) & 27.3±22.5 (\underline{51.9}) \\
\textbf{Radar}      & 4.2±0.5 (5.1)   & 18.0±5.2 (23.8)  & OOM\_G           \\
\textbf{ANOMALOUS}  & 3.4±2.3 (10.1)  & 13.6±4.6 (25.1)  & OOM\_G           \\
\midrule
\textbf{GCNAE}      & 18.8±0.0 (18.8) & 45.0±0.1 (45.1)  & 26.8±5.0 (28.1)  \\
\textbf{DOMINANT}   & 23.8±4.0 (26.1) & 19.4±0.9 (20.2)  & \textbf{38.9}±17.2 (48.4) \\
\textbf{DONE}       & \textbf{27.9}±9.8 (\underline{44.9}) & 20.6±9.4 (39.6)  & 23.6±2.3 (26.6)  \\
\textbf{AdONE}      & 22.0±5.9 (34.8) & 26.9±5.9 (38.2)  & 19.3±4.4 (27.0)  \\
\textbf{AnomalyDAE} & 21.6±1.9 (25.4) & 30.1±12.0 (44.8) & 17.7±7.7 (27.8)  \\
\textbf{GAAN}       & 19.6±0.3 (20.3) & \textbf{45.4}±0.1 (\underline{45.7})  & 28.0±0.1 (28.1)  \\
\textbf{GUIDE}      & 18.8±0.6 (20.3) & OOM\_C           & OOM\_C           \\
\textbf{CONAD}      & 19.8±7.3 (25.4) & 18.6±3.0 (20.2)  & 12.1±2.4 (14.3) \\
\bottomrule
\end{tabular}}
\vspace{-0.2in}
\end{table}
\vspace*{\fill}
\begin{table}[!ht]
\centering
\caption{\small Recall@k (\%) comparison among OD algorithms on six datasets with organic outliers, where we show \textit{the avg perf.} $\pm$ \textit{the STD of perf.} (\textit{max perf.}) of each. The best algorithm by expectation is shown in \textbf{bold}, while the max performance per dataset is marked with \underline{underline}. OOM denotes out of memory with regard to GPU (\_G) and CPU (\_C). TLE denotes time limit of 24 hours exceeded.}
\label{tab:main-recall-real}
\scalebox{0.74}{
\begin{tabular}{@{}l|cccccc@{}}
\toprule
\textbf{Algorithm} & \textbf{Weibo}  & \textbf{Reddit}   & \textbf{Disney}   & \textbf{Books}       & \textbf{Enron}     & \textbf{DGraph}      \\ 
\midrule
\textbf{LOF}        & 22.0±0.0 (22.0)  & \textbf{4.4}±0.0 (4.4)   & 0.0±0.0 (0.0)   & 0.0±0.0 (0.0)  & 0.0±0.0 (0.0) & TLE           \\
\textbf{IF}         & 13.8±6.4 (24.3)  & 0.1±0.1 (0.3)   & \textbf{9.2}±8.3 (16.7)  & 1.1±1.6 (3.6)  & 0.0±0.0 (0.0) & 0.1±0.1 (0.4) \\
\textbf{MLPAE}       & 48.9±11.0 (62.1) & 3.0±0.0 (3.0)   & 0.0±0.0 (0.0)   & 0.9±1.6 (3.6)  & 0.0±0.0 (0.0) & \textbf{0.5}±0.1 (\underline{0.6}) \\
\midrule
\textbf{SCAN}      & 23.8±7.0 (30.5)  & 2.7±0.3 (3.0)   & 7.5±11.2 (\underline{33.3}) & 0.7±1.4 (3.6)  & 0.0±0.0 (0.0) & TLE           \\
\textbf{Radar}      & \textbf{86.4}±0.8 (\underline{87.4})  & 2.1±0.8 (3.5)   & 0.0±0.0 (0.0)   & 0.0±0.0 (0.0)  & 0.0±0.0 (0.0) & OOM\_C        \\
\textbf{ANOMALOUS}  & \textbf{86.4}±0.8 (\underline{87.4})  & 4.0±1.9 (\underline{7.9})   & 0.0±0.0 (0.0)   & 0.0±0.0 (0.0)  & 0.0±0.0 (0.0) & OOM\_C        \\
\midrule
\textbf{GCNAE}      & 67.6±5.2 (77.3)  & 3.0±0.0 (3.0)   & 0.0±0.0 (0.0)   & 0.7±1.8 (7.1)  & 0.0±0.0 (0.0) & 0.4±0.0 (0.4) \\
\textbf{DOMINANT}   & 19.7±13.8 (37.4) & 0.9±0.4 (2.7)   & 3.3±6.7 (16.7)  & 1.6±3.1 (\underline{10.7}) & 0.0±0.0 (0.0) & OOM\_C        \\
\textbf{DONE}       & 65.4±12.4 (76.3) & 2.8±1.6 (5.7)   & 0.0±0.0 (0.0)   & 1.1±1.6 (3.6)  & 0.0±0.0 (0.0) & OOM\_C        \\
\textbf{AdONE}      & 64.3±7.6 (74.3)  & 1.0±1.2 (3.8)   & 1.7±5.0 (16.7)  & \textbf{3.0}±1.7 (7.1)  & 0.0±0.0 (0.0) & OOM\_C        \\
\textbf{AnomalyDAE} & 42.2±23.7 (75.7) & 0.9±0.5 (3.0)   & 0.0±0.0 (0.0)   & 2.7±2.2 (7.1)  & 0.0±0.0 (0.0) & OOM\_C        \\
\textbf{GAAN}       & 77.1±0.2 (77.4)  & 1.1±0.4 (2.2)   & 0.0±0.0 (0.0)   & 1.8±1.8 (3.6)  & 0.0±0.0 (0.0) & OOM\_C        \\
\textbf{GUIDE}      & OOM\_C           & OOM\_C          & 0.0±0.0 (0.0)   & 0.4±1.1 (3.6)  & OOM\_C        & OOM\_C        \\
\textbf{CONAD}      & 20.3±13.3 (37.1) & 1.3±1.6 (7.6)   & 0.8±3.6 (16.7)  & 1.7±2.9 (\underline{10.7}) & 0.0±0.0 (0.0) & 0.4±0.1 (\underline{0.6)}\\
\bottomrule
\end{tabular}}
\end{table}

\vspace*{\fill}
\clearpage
\newpage

\subsection{Additional Results on Performance Variation under Different Types of Outliers}
\vspace*{\fill}
\begin{table}[h]
\centering
\caption{\small Average Precision (\%) comparison among OD algorithms on three datasets injected with contextual and structural outliers, where we show \textit{the avg perf.} $\pm$ \textit{the STD of perf.} (\textit{max perf.}) of each. The best algorithm by expectation is shown in \textbf{bold}, while the max performance per dataset is marked with \underline{underline}. OOM denotes out of memory with regard to GPU (\_G) and CPU (\_C).}
\label{tab:type-ap}
\scalebox{0.74}{
\begin{tabular}{@{}l|cccccc@{}}
\toprule
\textbf{}           & \multicolumn{2}{c}{\textbf{Cora}}             & \multicolumn{2}{c}{\textbf{Amazon}}           & \multicolumn{2}{c}{\textbf{Flickr}}           \\
\cmidrule{2-7}
\textbf{Algorithm}           & Contextual            & Structural            & Contextual            & Structural            & Contextual            & Structural            \\ \midrule
\textbf{LOF}         & 12.2±0.0 (12.2)                & 3.1±0.0 (3.1)                  & 3.2±0.0 (3.2)                  & 2.6±0.0 (2.6)                  & 3.5±0.0 (3.5)                  & 2.5±0.0 (2.5)                  \\
\textbf{IF}          & 9.2±1.1 (11.7)                 & 3.0±0.3 (3.6)                  & 4.7±2.1 (10.5)                 & 2.6±0.1 (2.7)                  & 7.2±1.0 (9.1)                  & 2.6±0.0 (2.6)                  \\
\textbf{MLPAE}       & 13.7±0.0 (13.7)                & 3.1±0.0 (3.1)                  & 56.7±0.1 (\underline{56.9})                & 2.5±0.0 (2.5)                  & \textbf{24.8}±0.0 (\underline{24.9})                & 2.6±0.0 (2.6)                  \\ \midrule
\textbf{SCAN}       & 2.6±0.0 (2.7)   & 21.6±17.1 (61.0) & 2.5±0.0 (2.5)    & 22.4±24.4 (\underline{71.0}) & 2.5±0.0 (2.5)   & 59.1±36.8 (\underline{94.8}) \\
\textbf{Radar}      & 2.5±0.1 (2.6)   & 5.7±0.8 (6.6)    & 12.5±2.3 (14.3)  & 3.4±0.1 (3.6)    & OOM\_G          & OOM\_G           \\
\textbf{ANOMALOUS}  & 2.7±0.2 (3.4)   & 5.1±2.5 (13.5)   & 10.4±1.4 (13.7)  & 3.4±0.4 (4.0)    & OOM\_G          & OOM\_G           \\
\midrule
\textbf{GCNAE}      & 13.7±0.0 (13.7) & 3.1±0.0 (3.1)    & \textbf{56.8}±0.1 (\underline{56.9})  & 2.5±0.0 (2.5)    & 18.9±9.1 (\underline{24.9}) & 2.6±0.0 (2.6)    \\
\textbf{DOMINANT}   & 6.3±1.1 (6.9)   & 19.2±4.7 (22.1)  & 4.7±0.5 (5.2)    & 16.5±2.4 (17.4)  & 4.8±0.5 (5.1)   & \textbf{61.3}±13.7 (66.9) \\
\textbf{DONE}       & 7.0±2.2 (11.9)  & \textbf{31.9}±25.9 (\underline{89.0}) & 12.0±4.9 (20.6)  & \textbf{23.0}±18.1 (69.1) & 14.9±2.2 (18.0) & 9.8±2.0 (13.6)   \\
\textbf{AdONE}      & 8.9±1.3 (10.9)  & 16.2±6.2 (32.8)  & 14.5±7.4 (32.5)  & 13.1±8.3 (34.1)  & 11.4±2.6 (16.5) & 10.2±1.5 (13.0)  \\
\textbf{AnomalyDAE} & 9.3±1.7 (13.2)  & 14.8±4.9 (24.7)  & 21.9±17.7 (47.7) & 11.6±5.4 (17.2)  & 9.4±6.6 (24.2)  & 3.1±0.4 (3.8)    \\
\textbf{GAAN}       & \textbf{14.1}±0.1 (\underline{14.2}) & 4.5±0.1 (4.7)    & 52.5±1.4 (56.2)  & 3.6±0.2 (4.0)    & 24.7±0.2 (\underline{24.9}) & 2.6±0.0 (2.6)    \\
\textbf{GUIDE}      & 13.7±0.5 (14.0) & 3.9±0.2 (4.8)    & OOM\_C           & OOM\_C           & OOM\_C          & OOM\_C           \\
\textbf{CONAD}      & 6.5±1.1 (6.9)   & 20.0±4.6 (22.2)  & 4.8±0.5 (5.2)    & 16.9±1.0 (17.4)  & 4.5±0.3 (4.9)   & 5.4±0.3 (5.8)    \\
\bottomrule
\end{tabular}}
\end{table}
\vspace*{\fill}
\begin{table}[h]
\centering
\caption{\small Recall@k (\%) comparison among OD algorithms on three datasets injected with contextual and structural outliers, where we show \textit{the avg perf.} $\pm$ \textit{the STD of perf.} (\textit{max perf.}) of each. The best algorithm by expectation is shown in \textbf{bold}, while the max performance per dataset is marked with \underline{underline}. k is set as the number of each type of outliers in labels. OOM denotes out of memory with regard to GPU (\_G) and CPU (\_C).}
\label{tab:types-recall}
\scalebox{0.74}{
\begin{tabular}{@{}l|cccccc@{}}
\toprule
\textbf{}           & \multicolumn{2}{c}{\textbf{Cora}}             & \multicolumn{2}{c}{\textbf{Amazon}}           & \multicolumn{2}{c}{\textbf{Flickr}}           \\
\cmidrule{2-7}
\textbf{Algorithm}           & Contextual            & Structural            & Contextual            & Structural            & Contextual            & Structural            \\ \midrule
\textbf{LOF}        & 12.9±0.0 (12.9) & 5.7±0.0 (5.7)    & 1.1±0.0 (1.1)    & 3.7±0.0 (3.7)    & 9.0±0.0 (9.0)    & 2.2±0.0 (2.2)    \\
\textbf{IF}         & 13.1±3.3 (\underline{20.0}) & 3.8±2.0 (8.6)    & 4.4±3.5 (13.4)   & 2.1±0.7 (3.1)    & 13.8±1.9 (17.7)  & 2.5±0.3 (3.0)    \\
\textbf{MLPAE}      & 15.7±0.0 (15.7) & 7.1±0.0 (7.1)    & \textbf{55.1}±0.1 (55.1)  & 2.3±0.0 (2.3)    & \textbf{29.8}±0.1 (30.0)  & 2.9±0.0 (2.9)    \\
\midrule
\textbf{SCAN}       & 2.8±1.1 (4.3)   & 25.9±19.9 (61.4) & 2.2±0.4 (2.9)    & \textbf{25.1}±26.8 (70.3) & 2.8±0.2 (3.1)    & 59.7±37.1 (\underline{96.4}) \\
\textbf{Radar}      & 1.4±0.0 (1.4)   & 0.6±0.7 (1.4)    & 15.1±6.5 (20.6)  & 1.5±1.1 (2.3)    & OOM\_G           & OOM\_G           \\
\textbf{ANOMALOUS}  & 1.8±1.4 (4.3)   & 1.2±2.4 (8.6)    & 5.9±5.4 (19.1)   & 0.6±1.4 (4.6)    & OOM\_G           & OOM\_G           \\
\midrule
\textbf{GCNAE}      & 15.7±0.0 (15.7) & 7.1±0.0 (7.1)    & \textbf{55.1}±0.1 (55.1)  & 2.3±0.0 (2.3)    & 21.6±12.7 (30.1) & 2.8±0.2 (3.0)    \\
\textbf{DOMINANT}   & 8.4±3.1 (10.0)  & 17.2±5.2 (30.0)  & 4.9±0.8 (6.0)    & 10.1±1.9 (10.9)  & 3.6±0.2 (4.0)    & \textbf{70.6}±15.7 (76.0) \\
\textbf{DONE}       & 8.6±3.9 (18.6)  & \textbf{31.4}±26.2 (\underline{82.9}) & 11.6±4.0 (21.1)  & 23.7±19.9 (\underline{73.1}) & 22.4±3.0 (25.8)  & 5.1±1.0 (6.7)    \\
\textbf{AdONE}      & 11.4±3.1 (17.1) & 13.4±7.2 (35.7)  & 18.1±7.4 (30.6)  & 10.4±9.8 (30.9)  & 17.7±3.2 (23.9)  & 4.4±1.4 (7.7)    \\
\textbf{AnomalyDAE} & 12.0±2.9 (17.1) & 16.0±4.2 (22.9)  & 23.5±21.4 (53.4) & 8.8±2.9 (12.9)   & 10.1±10.5 (\underline{30.2}) & 3.6±1.7 (6.9)    \\
\textbf{GAAN}       & 15.9±0.4 (17.1) & 8.2±0.6 (8.6)    & 54.9±0.6 (\underline{56.3})  & 3.8±0.3 (4.6)    & \textbf{29.8}±0.1 (30.1)  & 2.9±0.0 (3.0)    \\
\textbf{GUIDE}      & \textbf{16.9}±0.6 (17.1) & 8.5±0.3 (8.6)    & OOM\_C           & OOM\_C           & OOM\_C           & OOM\_C           \\
\textbf{CONAD}      & 9.1±2.4 (10.0)  & 17.2±3.9 (18.6)  & 5.3±0.8 (6.0)    & 10.2±1.2 (10.9)  & 7.8±1.9 (10.1)   & 8.3±1.4 (9.7)    \\
\bottomrule
\end{tabular}}
\end{table}
\vspace*{\fill}
\clearpage
\newpage

\subsection{Additional Results on Efficiency and Scalability Analysis}
\label{appedix:time_mem}
\vspace*{\fill}
\begin{table}[h]
\centering
\caption{\small Time consumption (s) comparison among OD algorithms on five different numbers of epochs. For non-iterative algorithms, i.e., LOF, IF, and SCAN, we report the total runtime.}
\label{tab:time}

\footnotesize
\scalebox{0.9}{
\begin{tabular}{@{}l|lllll@{}}
\toprule
\textbf{Algorithm} & \textbf{10}         & \textbf{100}       & \textbf{200}       & \textbf{300}        & \textbf{400}       \\ \midrule
\textbf{LOF}        & 0.10 & 0.10 & 0.10 & 0.10 & 0.10      \\
\textbf{IF}         & 0.09 & 0.09 & 0.09 & 0.09 & 0.09             \\
\textbf{MLPAE}      & 0.04  & 0.46  & 0.82  & 1.37  & 1.74  \\
\midrule
\textbf{SCAN}       & 0.02  & 0.02  & 0.02  & 0.02  & 0.02              \\
\textbf{Radar}      & 0.02  & 0.10  & 0.19  & 0.34  & 0.36  \\
\textbf{ANOMALOUS}  & 0.02  & 0.09  & 0.17  & 0.26  & 0.36  \\
\midrule
\textbf{GCNAE}      & 0.06  & 0.49  & 0.96  & 1.45  & 1.94  \\
\textbf{DOMINANT}   & 0.08  & 0.70  & 1.41  & 2.10  & 2.79  \\
\textbf{DONE}       & 0.08  & 0.77  & 1.53  & 2.30  & 3.08  \\
\textbf{AdONE}      & 0.10  & 0.91  & 1.81  & 2.71  & 3.62  \\
\textbf{AnomalyDAE} & 0.43  & 0.64  & 1.28  & 1.92  & 2.55  \\
\textbf{GAAN}       & 0.06  & 0.49  & 0.98  & 1.47  & 1.97  \\
\textbf{GUIDE}      & 50.77 & 51.92 & 53.40 & 54.27 & 55.21 \\
\textbf{CONAD}      & 0.11  & 1.04  & 2.07  & 3.07  & 4.10  \\
\bottomrule
\end{tabular}}
\vspace{-0.2in}
\end{table}
\vspace*{\fill}
\begin{table}[h]
\centering
\caption{\small GPU memory consumption (MB) comparison among deep algorithms on five different graph sizes (number of nodes). Note that GPU memory measurement does not apply to algorithms like LOF, IF, and SCAN.}
\label{tab:mem}
\footnotesize
\scalebox{0.9}{
\begin{tabular}{@{}l|ccccc@{}}
\toprule
\textbf{Algorithm} & \textbf{100}         & \textbf{500}       & \textbf{1000}       & \textbf{5000}        & \textbf{10000}       \\ \midrule
\textbf{MLPAE}      & 0.66         & 2.10         & 3.90          & 19.41         & 38.52          \\
\textbf{GCNAE}      & 0.92         & 3.40         & 6.38          & 31.18         & 62.11          \\
\textbf{GUIDE}      & 0.96         & 3.46         & 6.47          & 31.45         & 62.60          \\
\textbf{Radar}      & 0.49         & 9.32         & 36.73         & 871.32        & 3450.88        \\
\textbf{ANOMALOUS}  & 0.44         & 5.03         & 20.50         & 482.47        & 2293.76        \\
\textbf{DOMINANT}   & 1.09         & 7.58         & 27.15         & 591.74        & 2324.48        \\
\textbf{DONE}       & 1.95         & 10.93        & 32.74         & 624.41        & 2385.92        \\
\textbf{AdONE}      & 1.99         & 11.38        & 33.60         & 657.54        & 2447.36        \\
\textbf{AnomalyDAE} & 1.21         & 10.54        & 36.94         & 794.81        & 3112.96        \\
\textbf{GAAN}       & 0.90         & 9.51         & 36.87         & 871.17        & 3450.88        \\
\textbf{CONAD}      & 1.39         & 8.77         & 29.48         & 604.32        & 2344.96        \\
\bottomrule
\end{tabular}}
\vspace{-0.2in}
\end{table}
\vspace*{\fill}
\clearpage
\newpage


\newcommand{\system}{{PyGOD}\xspace}

\section{\rv{Long-term Maintenance and Development Plan}}

\rv{We commit to maintaining and developing \method and \system in the long run, as many of our open-source outlier detection works (e.g., PyOD \cite{zhao2019pyod}, SUOD \cite{zhao2021suod}, and TODS \cite{lai2021tods}). More specifically, we will focus on improving on two aspects of graph OD tasks, namely datasets (Appx. \ref{appx:new_datasets}) and algorithms (Appx. \ref{appx:new_algorithms})}

\subsection{\rv{Enriching Graph OD Datasets}}
\label{appx:new_datasets}
\rv{We will keep monitoring the coming datasets suited for \method tasks, and enrich our testbed with more datasets. There are three main approaches for this:
\setlist{nolistsep}
\begin{enumerate}[leftmargin=*,noitemsep]
\item \textbf{\textit{Directly including new graph OD datasets}}. We will keep checking graph OD papers to include their newly introduced datasets.
\item \textit{\textbf{Adapting graph datasets for graph OD tasks}}. As we have discussed in future directions in \S \ref{sec06:con}, we could repurpose existing graph datasets for OD. For instance, given a graph dataset with multiple types of transactions for node classification, we could combine the rare classes together as anomalies, and the common transactions as the normal class. Most of the time, we could find some semantic meaning for the combined rare classes, e.g., fraud and mistakes. This adaptation process has been widely used in tabular OD \cite{emmott2015meta} and has proven to be useful \cite{campos2016evaluation}. Specifically, Open Graph Benchmark (OGB) \cite{ogb}, and therapeutic data commons (TDC) \cite{huang2021therapeutics} can serve as natural sources for building graph OD datasets, and we will start from these repositories.
\item\textbf{ \textit{Planting more types of synthesized outliers into plain graphs.}} Our experimental results and analysis suggest that the existing synthesizing approaches are too naive and not similar to most organic outliers. Our future plan includes: 1) adopting other outlier generation approaches from~\cite{xu2007scan, bandyopadhyay2019outlier};
2) generating outliers using learning-based methods like GAN~\cite{steinbuss2021benchmarking}.
\end{enumerate}}

\rv{With more graph OD datasets (e.g., \# datasets $>=20$), we could conduct more in-depth (group-wise and pairwise) statistical analysis \cite{demvsar2006statistical}, which has not been possible in \method works. We will keep updating the benchmark site\footnote{\url{https://github.com/pygod-team/pygod/tree/main/benchmark}} for newly added datasets.}

\subsection{\rv{Emerging Graph OD Algorithms}}
\label{appx:new_algorithms}

\rv{Regarding graph OD algorithms, we will keep maintaining and improving \system in multiple aspects:
\begin{enumerate}[leftmargin=*,noitemsep]
\item \textbf{\textit{Monitoring and adding outlier node methods to \system}} for both benchmark and general usage.
\item \textbf{\textit{Optimizing its accessibility and scalability}} with the latest development in graph learning \cite{jia2020improving}, which may bring us new insights into outlier node detection's scalability.
\item \textbf{\textit{Incorporating automated machine learning}} to enable intelligent model selection and hyperparameter tuning~\cite{park2022autogml,zhao2021automatic}, which may unlock some interesting perspectives of graph OD.
\item \textbf{\textit{Extending the scope from static attribute outlier node detection to more graph tasks}}, e.g., outlier detection in edges and sub-graphs. This will lead to other interesting aspects of graph OD. 
\end{enumerate}}

\rv{\textbf{Robustness and Quality}. While building \system, we follow the best practices of system design and software development.
First, we leverage the continuous integration by \textit{GitHub Actions}\footnote{Continuous integration by GitHub Actions: \url{https://github.com/pygod-team/pygod/actions}} to automate the testing process under various Python versions and operating systems. 
In addition to the scheduled daily test, both commits and pull requests trigger the unit testing.
Notably, we enforce all code to have at least 90\% coverage\footnote{Code coverage by Coveralls: \url{https://coveralls.io/github/pygod-team/pygod}}.
Following the \texttt{PEP8} standard, we enforce a consistent coding style and naming convention, which facilitates community collaboration and code readability.}

\rv{In the long term, we envision \system could keep evolving to support more comprehensive benchmarking, as well as other graph detection tasks and benchmarks.}



\end{document}